%% file: main.tex
\documentclass[runningheads]{llncs}

\usepackage{eccv}

\usepackage{eccvabbrv}

\usepackage{graphicx}
\usepackage{booktabs}

\usepackage[accsupp]{axessibility}  

\usepackage{hyperref}

\usepackage{orcidlink}

\usepackage{enumitem}
\usepackage[marginal]{footmisc}

\usepackage{url}
\usepackage{capt-of}
\makeatletter
\newcommand{\@chapapp}{Appendix}
\makeatother
\usepackage{appendix}

\begin{document}

\title{Kinema4D: Kinematic 4D World Modeling for Spatiotemporal Embodied Simulation} 

\titlerunning{Kinema4D}

\footnotetext[2]{\label{footnote:corrs}Corresponding author.}
\author{%
Mutian Xu\inst{1} \quad Tianbao Zhang\inst{2} \\
\quad Tianqi Liu\inst{1} \quad Zhaoxi Chen\inst{1} \quad Xiaoguang Han\inst{2} \quad Ziwei Liu\inst{1}$^{\ref{footnote:corrs}}$
}

\authorrunning{Xu et al.}

\institute{S-Lab, Nanyang Technological University \and
SSE, CUHKSZ \vspace{5pt}\\
\small{\href{https://mutianxu.github.io/Kinema4D-project-page/}{mutianxu.github.io/Kinema4D-project-page}}
}

\maketitle

\begin{figure}[h]
\centering
\includegraphics[width=0.9\textwidth]{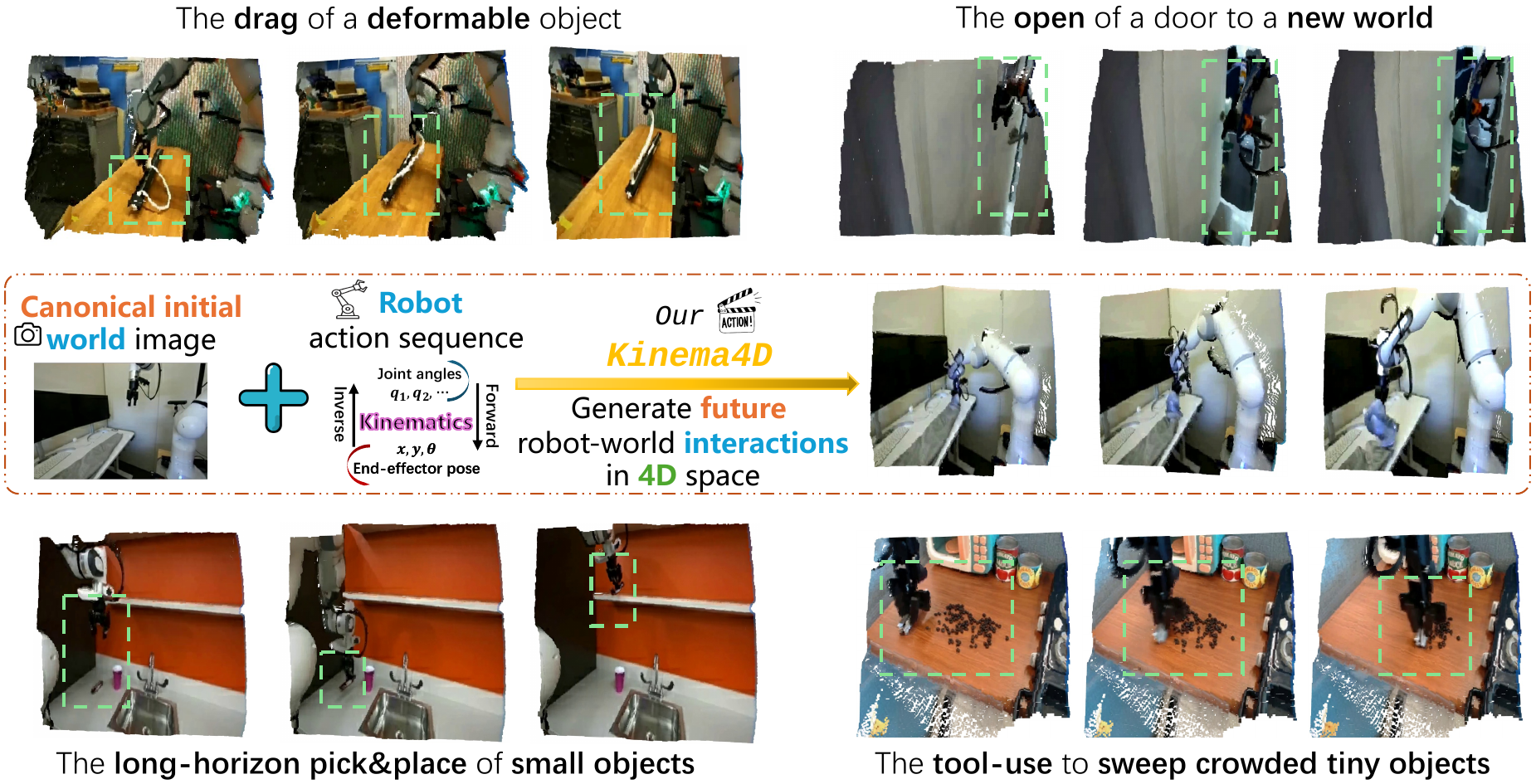}
\caption{We propose \textbf{Kinema4D}, a new \textit{action-conditioned 4D generative embodied simulator}. Given an initial world image with a robot at a canonical setup space, and an action sequence, our method generates future robot-world interactions in 4D space. It simulates physically plausible and geometrically consistent interactions between complex robot actions and diverse objects across various spatial constraints, providing a new foundation for advancing next-generation embodied simulation.
}
\label{fig:teaser}
\vspace{-0.4cm}
\end{figure}

\input{sec/0_abstract}    
\input{sec/1_intro}
\input{sec/2_related_work}
\input{sec/3_method}
\input{sec/4_experiment}
\input{sec/5_conclusion}

\clearpage  


%
%

\input{main.bbl}
\bibliographystyle{splncs04}

\input{sec/x_supple}

\end{document}

%% file: sec/0_abstract.tex
\begin{abstract}
Simulating robot-world interactions is a cornerstone of Embodied AI. Recently, a few works have shown promise in leveraging video generations to transcend the rigid visual/physical constraints of traditional simulators.
However, they primarily operate in 2D space or are guided by static environmental cues, ignoring the fundamental reality that robot-world interactions are \textit{inherently 4D spatiotemporal events} that require \textit{precise} interactive modeling.
To restore this 4D essence while ensuring the precise robot control,
we introduce Kinema4D, a new action-conditioned 4D generative robotic simulator that disentangles the robot-world interaction into:
i) \textit{Precise} 4D representation of \textit{robot controls}: we drive a URDF-based 3D robot via kinematics, producing a precise 4D robot control trajectory. ii) \textit{Generative} 4D modeling of \textit{environmental reactions}: we project the 4D robot trajectory into a pointmap as a spatiotemporal visual signal, controlling the generative model to synthesize complex environments’ reactive dynamics into synchronized RGB/pointmap sequences.
To facilitate training, we curated a large-scale dataset called Robo4D-200k, comprising 201,426 robot interaction episodes with high-quality 4D annotations. Extensive experiments demonstrate that our method effectively simulates physically-plausible, geometry-consistent, and embodiment-agnostic interactions that faithfully mirror diverse real-world dynamics. For the first time, it shows potential \textit{zero-shot} transfer capability, providing a high-fidelity foundation for advancing next-generation embodied simulation.

\end{abstract}

%% file: sec/1_intro.tex
\section{Introduction}
In the field of Embodied AI, the ability to roll out robot trajectories within a world environment is pivotal for scaling up robot demonstrations \cite{o2024open}, policy evaluation \cite{li2024evaluating,team2025evaluating}, and reinforcement learning \cite{ebert2018visual}. Nevertheless, executing actions in the real world is prohibitively costly, potentially unsafe, and necessitates constant expert maintenance \cite{brohan2022rt,khazatsky2024droid}. Consequently, robotic simulation in virtual environments has emerged as a vital surrogate for real-world deployment.

While significant efforts have been dedicated to developing robust physical simulators \cite{xiang2020sapien,makoviychuk2021isaac,mittal2023orbit,chen2024urdformer}, these platforms often lack visual realism and rely on hand-crafted and pre-defined physical properties/rules. Such dependencies create a significant scalability bottleneck, particularly when attempting to synthesize new environments. 
Very recently, to circumvent these limitations, researchers have begun leveraging the intrinsic world dynamics captured by video generative models to synthesize robot-environment interactions across diverse scenes \cite{zhu2025irasim,agarwal2025cosmos}.
By casting robot actions as conditional prompts for video synthesis, these methods directly simulate the visual outcome of adding robot controls to an environment, bypassing the need for predefined physical modeling required for traditional simulation.
However, a critical gap remains. Current models primarily operate within a 2D pixel space \cite{zhu2025irasim,agarwal2025cosmos}, whereas robot-world interactions are \textit{inherently 4D spatiotemporal events}. Without 4D constraints, physical interaction loses its most fundamental grounding. Although a few latest works  \cite{zhen2025tesseract,robo4dgen} explored 4D simulation, they primarily rely on high-level linguistic instructions to represent robot control. Such semantic representations, while intuitive, \textit{lack the precise guidance} essential for high-fidelity 4D world modeling.
As a result, existing methods struggle to simulate precise interactive effects—such as material deformations or occluded object dynamics (see \cref{fig:qualitative_result_2D,fig:qualitative_result_4D}).

To bridge this gap and advance this problem, we propose Kinema4D, a new action-conditioned 4D generative robotic simulator. Our core motivation is to restore robot-world interactions to their \textit{4D spatiotemporal essence}, while ensuring \textit{precise robot control}.
This motivation is further grounded in two synergetic insights that support our design philosophy—disentangling the simulation into \textit{robot control} and their \textit{resultant environmental changes}:

\textbf{i)} \textbf{\textit{Precise} 4D representation of robot actions via kinematic control}: We recognize the fact that robot action is a \textit{precise physical certainty in 4D space} and should not be “guessed” by a generative model. A joint-angle sequence or an end-effector pose is merely an abstract vector until it is mapped onto a physical structure. Reflecting on this, our framework enables a bridge that is otherwise impossible in 2D: we drive a 3D-reconstructed, URDF-based robot model via explicit kinematics to produce a continuous 4D robot trajectory that is guaranteed to be kinematically correct, providing the high-granularity spatiotemporal information that serves as the \textit{causal driver} for any interaction.
\textbf{ii)} \textbf{\textit{Generative} 4D modeling of environmental reactions via controllable generation}: While robot controls are deterministic, we argue that complex environments’ dynamics require a \textit{flexible generative modeling}. To this end, we project the previously derived 4D robot trajectory into a pointmap sequence—serving as a spatiotemporal visual signal, which controls the generative model to offload the burden of kinematic modeling and focus exclusively on synthesizing the environment's \textit{reactive dynamics}.
Moreover, our architecture simultaneously predicts synchronized RGB and pointmap sequences, effectively transforming the generation into a spatiotemporal reasoning task within a unified 4D space. This makes the result not only visually realistic but also geometrically consistent.
In this paradigm, spatiotemporal awareness is enforced not merely as a control signal but as an intrinsic constraint throughout the generative process. This fosters a symbiosis between \textit{precise control} and \textit{flexible synthesis}.

To facilitate the training of our framework, we curated a large-scale dataset called Robo4D-200k, comprising 201,426 real-world and synthetic demonstrations with high-quality 4D annotations.
We extensively benchmark our method using both video and geometric metrics, as well as policy evaluation. 
Experiments show that our Kinema4D generalizes to simulate physically-plausible robot-world interactions that closely mirror diverse real-world dynamics, and for the first time, shows potential zero-shot OOD transfer capability.

Our contributions are summarized as:
\begin{itemize}
[itemsep=2pt,topsep=2pt,parsep=0pt, leftmargin=2em]
    \item Kinema4D, a new action-conditioned 4D generative robotic simulator that enjoys both spatiotemporal rigor and generative flexibility.
    \item Robo4D-200k, to our knowledge, the largest-scale 4D robotic dataset comprising 201,426 demonstrations with high-quality 4D annotations. 
    \item A comprehensive evaluation scheme demonstrating that our method mirrors real-world dynamics, providing a new foundation for embodied simulation. 
    \item Our code, dataset, and checkpoints will all be public upon acceptance.
\end{itemize}


%% file: sec/2_related_work.tex
\section{Related Work}
\label{sec:related_work}
We review the evolution of embodied simulation, tracing its trajectory from classical physics engines to the recent emergence of generative world models.

\paragraph{Physical simulation.}
Classical robotic simulation methods rely on physics engines, such as MuJoCo, IssacSim, SAPIEN \cite{todorov2012mujoco,erez2015simulation,xiang2020sapien,makoviychuk2021isaac,mittal2023orbit,robotwin}. They simulate interactions based on rigid-body dynamics, requiring meticulously hand-crafted meshes, precise physical properties (\eg, friction, mass), and pre-defined rules. 
Recent real-to-sim advancements further leverage 3D Gaussian Splatting (3DGS) with continuum mechanics or rigid-body solvers to reconstruct digital twins directly from real-world captures \cite{pashevich2019learning,li2024robogsim,abouchakra-embodiedgaussians,jiang2025gsworld,han2025re}. 
A few methods also focus on simulating realistic depth \cite{xu2025stable,liu2025manipulation}.
Nevertheless, they rely on explicit physical solvers and predefined laws, which struggle to generalize to complex environmental responses, limiting their scalability across unstructured environments.

\paragraph{Learning world models.}
World Models \cite{allen1983planning,ha2018world} aim to internalize the underlying dynamics of the environment, allowing agents to “dream” and plan within a learned latent space. With the advent of diffusion models \cite{song2019generative,ho2020denoising,dit} and large-scale video pre-training \cite{hong2022cogvideo,bruce2024genie,zhang2025show,wan2025wan}, video generative models have demonstrated an extraordinary ability to capture complex visual dynamics \cite{assran2025v,chefer2025videojam}. Furthermore, \textit{interactive} video generation has advanced through the integration of various conditioning mechanisms \cite{zhang2023adding} designed to represent diverse control signals. Early efforts in this direction have focused on injecting viewpoint trajectories \cite{yu2025wonderworld,liang2025wonderland,ren2025gen3c,gu2025diffusion,Yu_2025_ICCV} to synthesize immersive, navigable environments.

\paragraph{Embodied video-generation models.}
Recently, research has pivoted toward using embodied actions as controls for interactive video generation, aiming to synthesize robot-world interactions with both visual and physical fidelity. Based on the representation of embodied actions at the input level, this research stream can be categorized into four sub-groups.
1) Text: A main-stream works \cite{robodreamer,yang2024learning,robomaster,chen2025world4omni,yang2025roboenvision} generate visual outcomes from high-level text instructions, which lack the fine-grained precision required for low-level manipulations.
2) Latent embedding: Another type of works \cite{wu2024ivideogpt,cen2025worldvla,cen2025rynnvla,huang2025ladi,luo2025learning,zhang2025imowm,zhu2025irasim,agarwal2025cosmos,gao2026dreamdojo,liao2026genie,guo2026ctrl} such as IRASim \cite{zhu2025irasim} and Ctrl-World \cite{guo2026ctrl} typically encode 7-DoF end-effector poses into compressed embeddings. This forces the generative model to infer or `guess' the underlying robot kinematics, often resulting in physically implausible failures.
3) Semantics: ORV \cite{yang2026orv} introduces 3D semantic occupancy of static environments; however, it needs an off-the-shelf occupancy-prediction model to acquire and lacks temporal dynamics, still requiring action encodings or text to provide dynamic information.
4) 2D visual prompt: EVAC \cite{jiang2025enerverse}, AnchorDream \cite{ye2025anchordream}, and VAP \cite{VAP_2025_ICCV} represent embodied actions using 2D prompts like angle arrows, 2D renders, and skeletons. A concurrent work, BridgeV2W \cite{chen2026bridgev2wbridgingvideogeneration}, utilizes URDF to drive robots but merely renders the resulting trajectory into 2D binary masks. Such 2D-based signals lack spatiotemporal constraints and struggle to provide precise control guidance.
Commonly, as for the output, all aforementioned methods use video generations to predict 2D (\ie, RGB) frames, treating the robot/environment as a monolithic pixel stream. Due to the lack of spatiotemporal awareness, they struggle to simulate complex physical interactions. 

\textbf{Summary}: Whether evaluated by action conditioning or output modality, the aforementioned methods fail to resolve the \textit{\textbf{challenging trilemma}} of \textit{dynamics, precision} and \textit{spatiotemporal awareness}—the three pillars of reliable embodied simulation. To overcome this, our Kinema4D learns intricate dynamics through a 4D generative model, where abstract action controls are grounded via kinematics, securing precision and spatiotemporal awareness, which in turn anchors the generative model to synthesize complex dynamics.

\paragraph{3D/4D world models.}
Several recent works have been able to output 3D/4D worlds. First, ParticleFormer \cite{huang2025particleformer} and PointWorld \cite{huang2026pointworld} directly utilize simple transformers to predict the trajectories of pre-defined 3D particles for both robots and environments. However, they lack the generative flexibility required to synthesize new geometry that depicts emergent world dynamics beyond the initial 3D input.
On the other hand, 4D-native generative models that aim to generate full spatio-temporal sequences have gained significant attention. For instance, Aether \cite{Aether} unifies geometry-aware reasoning by jointly optimizing 4D dynamic reconstruction and goal-conditioned video prediction, while 4DNex \cite{chen20254dnex} enables 4D world generation in a single feed-forward pass. 
Although recent advancements such as TesserAct \cite{zhen2025tesseract}, GWM \cite{lu2025gwm}, iMoWM \cite{zhang2025imowm}, and Robo4DGen \cite{robo4dgen} demonstrate the ability to synthesize 4D embodied worlds at the output stage, they remain constrained by a critical design bottleneck: they solely inject \textit{static} guidance (\eg, depth maps, surface normals, or Gaussian splats) from the initial world environment, which \textit{lacks temporal dynamics}. Consequently, these models still rely on text instructions or latent tokens to inject robot actions, which \textit{lack the granularity} required for fine-grained interactions. In contrast, by transforming abstract robot actions into 4D pointmaps, our Kinema4D accepts \textit{precise} robot controls with \textit{both spatial and temporal cues}. To fully utilize such controls, we apply a 4D-native generative model with spatiotemporal reasoning abilities, allowing the model to not only understand robot movement but also to flexibly predict its physical effects on the environment for high-fidelity simulation.

%% file: sec/3_method.tex
\section{Our Approach}

\begin{figure}[t]
\centering
\includegraphics[width=\textwidth]{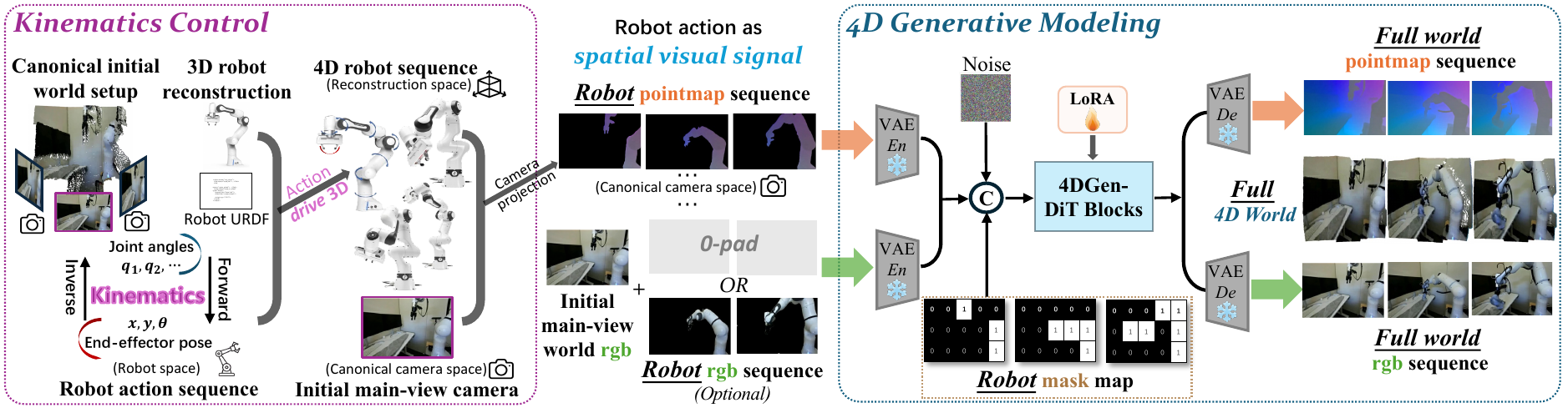}
\vspace{-0.4cm}
\caption{\textbf{Overview of our \textit{Kinema4D}}.
\textbf{\textit{1) Kinematics Control}}: Given a 3D robot with its URDF at initial canonical setup space, and an action sequence, we drive the 3D robot via kinematics to produce a 4D robot trajectory, which is then projected into a pointmap sequence. This process re-represents raw actions as a spatiotemporal visual signal.
\textbf{\textit{2) 4D Generative Modeling}}: This signal and the initial main-view world image are sent to a shared VAE encoder, then fused with an occupancy-aligned robot mask and noise, which are denoised by a Diffusion Transformer \cite{dit} to generate a full future 4D (pointmap+RGB) world sequence.}
\label{fig:pipeline}
\vspace{-0.4cm}
\end{figure}

\cref{fig:pipeline} illustrates our architecture, composed of two components: Kinematic Control (\cref{sec:kinematics_control}) and 4D Generative Modeling (\cref{sec:4d_gen_modeling}).

\subsection{Kinematics Control}
\label{sec:kinematics_control}
The core of our framework lies in transforming abstract robot actions into a precise 4D representation, via a progressive process.

\paragraph{3D robot asset acquisition.}
To ground the robot in 4D space, we first establish its geometric entity. For standardized robots, we utilize factory-provided 3D CAD meshes. For unknown platforms, we implement a reconstruction pipeline: 

We first capture orbital videos of the robot and sample representative frames, then employ Grounded-SAM2 \cite{ravi2024sam2, liu2023grounding} to segment the robot in the initial frame using a robust prompt (\eg, \texttt{“the robot arm”}), followed by SAM2 \cite{ravi2024sam2} to propagate masks across the sequence via video tracking. Next, given the masked multi-view images, we leverage ReconViaGen \cite{chang2025reconviagen} to recover a high-quality textured robot mesh $\mathcal{C}_{recon}$ in under one minute. This pipeline facilitates the efficient acquisition of our 3D asset library for diverse robotic configurations.
To enable articulation, we establish a digital twin alignment: the joint anchor points from the robot's URDF model $\mathcal{M}$ are mapped to their corresponding coordinates of $\mathcal{C}_{recon}$ within the canonical reconstruction space. This ensures that the analytical kinematic chain can directly drive the reconstructed mesh segments.

\vspace{-0.2cm}
\paragraph{Kinematics-driven 4D robot trajectory expansion.}
Given the aligned robot model $\mathcal{M}$ within $\mathcal{C}_{recon}$, we transform input actions $\mathbf{a}_{1:T}$ into full-body 4D trajectories. Our framework handles two primary control modalities:

\begin{itemize}[leftmargin=*, nosep]
    \item \textit{End-effector control}. When actions are provided as Cartesian poses $\{\mathbf{T}_{ee, t}\}_{t=1}^T$, we employ an Inverse Kinematics (IK) solver to resolve the joint configurations at time $t$: $\mathbf{q}_t = \text{IK}(\mathbf{T}_{ee, t}, \mathbf{q}_{t-1}, \mathcal{M})$, where the previous state $\mathbf{q}_{t-1}$ serves as the seed to ensure temporal smoothness and prevent joint-space flipping.
    \item \textit{Joint-space control}. When actions are directly provided as joint angles or velocities, $\mathbf{q}_t$ is obtained via direct mapping or integration.
\end{itemize}

For each time $t$, we perform Forward Kinematics (FK) to compute the 6-DoF poses for all $K$ links within the reconstruction space: $\{\mathbf{T}_{k,t}^{recon}\}_{k=1}^K = \text{FK}(\mathbf{q}_t, \mathcal{M})$.

\paragraph{Spatial-visual projection.}
We select a primary viewpoint—typically a medial-frontal perspective—for projection and later generation, as single-view robot demonstration data is extensively accessible to ensure training scalability and diversity. 
To place the articulated robot within the initial main-view world image $I_0$, we utilize the extrinsic camera transformation $\mathbf{T}_{recon}^{cam} \in SE(3)$ of the selected main view, which is derived during the previous 3D robot reconstruction phase \cite{chang2025reconviagen}.
With the camera transformation $\mathbf{T}_{recon}^{cam}$ and the previously computed link poses $\{\mathbf{T}_{k,t}^{recon}\}$ within the reconstruction space, the full-body trajectory is projected onto the image plane to generate the 4D robot pointmap $\mathbf{M}_{1:T} \in \mathbb{R}^{H \times W \times 3}$. For any point $\mathbf{x}$ on the surface of link $k$, its projected pixel coordinates $(u, v)$ and depth $z$ are determined by: 
{\scriptsize
\begin{equation}
\begin{bmatrix} u \cdot z \\ v \cdot z \\ z \end{bmatrix} = \mathbf{K} \cdot \mathbf{T}_{recon}^{cam} \cdot \mathbf{T}_{k,t}^{recon} \cdot \mathbf{x},
\end{equation}
} 
where $\mathbf{K}$ is the camera intrinsic matrix. The resulting pointmap $\mathbf{M}_{1:T}$ is pixel-aligned with the RGB grid, while its pixel values store the camera-space $(x, y, z)$ coordinates. This transformation precisely maps the 4D robot trajectory from the canonical reconstruction space into the target camera coordinate system, ensuring that the robot occupancy is spatially consistent with the environmental background. 
Optionally, we project the textured trajectory to RGB sequence. 

\cref{tab:ablation} ablates the robustness to the noisy pointmap. More details of our reconstruction and projection are provided in the supplementary material.

\subsection{4D Generative Modeling}
\label{sec:4d_gen_modeling}
Next, we utilize a 4D diffusion model to focus on synthesizing the environment’s reactive dynamics in response to the robot control signal.

\paragraph{Preliminary: Latent Video Diffusion.}
We build our framework upon Latent Diffusion Models (LDM) \cite{rombach2022high}, extending them to the 4D spatiotemporal domain. Instead of pixel-space denoising, the model operates on a compressed latent space $\mathcal{Z}$ provided by a pre-trained VAE. A video sequence $\mathbf{V}_{1:T}$ is encoded into a latent tensor $\mathbf{z}_0 \in \mathbb{R}^{T \times C \times H \times W}$. The diffusion process learns to generate this latent sequence by optimizing a conditional denoising objective:
{\scriptsize
\begin{equation}
    \mathcal{L}_{vid} = \mathbb{E}_{\mathbf{z}_0, \epsilon, \tau, \mathbf{c}} \left[ \| \epsilon - \epsilon_\theta(\mathbf{z}_\tau, \tau, \mathbf{c}) \|^2 \right],
\end{equation}
}
where $\mathbf{z}_\tau$ is the noisy latent at diffusion step $\tau$, and $\epsilon_\theta$ is typically a Spatio-Temporal Transformer (\eg, DiT \cite{dit}) that models dependencies across both frames and pixels. The condition $\mathbf{c}$ guides the denoising process to ensure the generated video adheres to specific inputs.

\paragraph{Multi-modal latent construction.}
To prepare the multi-modal input for our generative backbone, we first align the temporal dimensions of the initially captured RGB world image $I_0$ and the robotic control signals. Specifically, $I_0$ is extended along the temporal axis via zero-padding or by concatenating the previously robot RGB sequence.
Next, following the data formatting strategy in 4DNex \cite{chen20254dnex}, we concatenate this input with the previously produced robot pointmap sequence $\mathbf{M}_{1:T}^{robot}$ along the width dimension. This unified spatiotemporal signal is then processed by a shared VAE to obtain input latents, which maps the heterogeneous inputs—RGB and pointmap—into a synchronized latent representation.

To further enforce pixel-level control, we introduce a guided mask $\mathbf{m} \in \{0, 1\}^{T \times H \times W}$, where $\mathbf{m}_{t,i,j} = 1$ indicates the spatial occupancy of the robot (derived from $\mathbf{M}_{1:T}^{robot}$) and $0$ indicates the area to be generated. In practice, rather than employing a binary hard-mask, we implement a soft strategy by setting the value of 10\% occupied regions to $0.5$ (ablated in \cref{tab:ablation}). This design choice ensures that the generative model retains the capacity to \textit{refine} the robot's visual signal. By doing so, we mitigate the impact of noise introduced during the previous phase, enhancing the structural robustness and visual coherence.

We concatenate input latents, noisy latents, and robot masks channel-wise. This fused input ensures that the generative model focuses exclusively on synthesizing the environment's reactive dynamics while \textit{strictly adhering to the robot’s trajectory}.

\paragraph{4D-aware joint modeling.}
Our backbone is a Diffusion Transformer \cite{dit} that predicts synchronized RGB and pointmap sequences. To preserve pixel-wise alignment across these modalities, we adopt a shared Rotary Positional Encoding (RoPE) \cite{su2024roformer} across both RGB and pointmap latents. Additionally, following 4DNex \cite{chen20254dnex}, we distinguish these domains using learnable domain embeddings. These embeddings act as modality-specific signatures, allowing the Transformer to perform cross-modal reasoning—effectively using the robot pointmap as geometric anchors to guide the synthesis of the RGB environmental response.

\paragraph{4D sequence synthesis.} 
The denoised latents are processed by the shared VAE Decoder to reconstruct the full-world pointmap/RGB sequence. By predicting the environment's pointmap $\mathbf{M}_{1:T}^{world}$, the model produces a simulation that is not only visually realistic but also geometrically rigorous, ultimately yielding a 4D world where every pixel's depth and motion are grounded in 3D space.

\paragraph{Underlying insight.} 
By jointly modeling RGB and pointmap latents, the generative process is transformed into a spatiotemporal reasoning task. The model does not merely “draw” pixels; it must resolve the 3D occupancy and deformations of the world that are consistent with the robot's movement. 

\subsection{Robo4D-200k: A Large-Scale 4D Robotic Dataset}

\begin{figure}[t]
\centering
\includegraphics[width=0.9\textwidth]{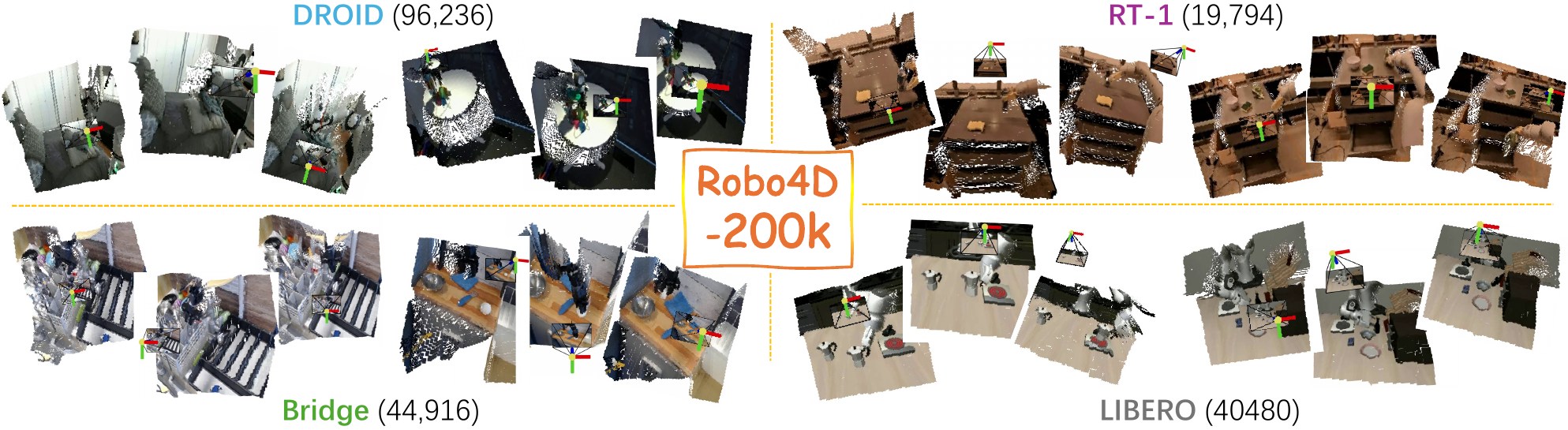}
\vspace{-0.2cm}
\caption{\textbf{Samples from our \textit{Robo4D-200k} dataset}.
Our dataset provides a comprehensive data foundation by aggregating diverse real-world demonstrations, including DROID \cite{khazatsky2024droid}, Bridge \cite{walke2023bridgedata}, and RT-1 \cite{rt12022arxiv}. We further incorporate the LIBERO \cite{libero} to synthesize a vast array of successful/failure cases. Each episode captures a \textit{complete} robot-world interaction (\eg, pick-and-place)—providing the continuous information necessary for robust reasoning. The 4D point clouds viewed from various camera frustums are shown here, demonstrating the spatial precision of our pseudo-annotations.}
\label{fig:dataset}
\vspace{-0.3cm}
\end{figure}

\paragraph{Data preparation.}
To facilitate the training of Kinema4D, a large-scale dataset is a prerequisite.
To begin with, we aggregate 2D RGB video streams from leading real-world robotic demonstration repositories, including DROID \cite{khazatsky2024droid}, Bridge \cite{walke2023bridgedata}, and RT-1 \cite{rt12022arxiv}. We further incorporate the LIBERO \cite{libero} platform to synthesize an extensive array of both successful executions and critical failure modes.

\paragraph{4D annotation.}
A significant challenge lies in lifting the raw 2D RGB videos from real-world datasets into the 4D metric space required for our joint generative modeling.
We extensively tried several state-of-the-art 3D/4D reconstruction frameworks, including  MonST3R \cite{zhang2024monst3r}, MegaSaM \cite{li2025megasam}, VGGT/VGGT4D \cite{wang2025vggt, hu2025vggt4d}, ViPE \cite{huang2025vipe}, and ST-V2 \cite{xiao2025spatialtracker}. ST-V2 \cite{xiao2025spatialtracker} yields the most robust and temporally consistent pointmap sequences, particularly for robot manipulations involving rapid motion. It is used to reconstruct high-quality, pixel-aligned 4D trajectories for real datasets. For synthetic data from LIBERO, we directly leverage the native, noise-free depth parameters to ensure absolute ground-truth precision.

\paragraph{Dataset curation.}
To ensure the quality of the generative simulation, we performed a rigorous data curation process. We applied a manual verification to prune low-quality captures and reconstruction artifacts. For very-long demonstrations, we applied uniform temporal downsampling to yield 49-frame sequences per episode; this ensures consistent motion frequencies across the dataset.
Each curated episode captures a complete/contiguous spatiotemporal interaction (\eg, a full pick-and-place cycle), providing the long-term dependencies necessary for robust physical reasoning.
Following this pipeline, we established \textit{Robo4D-200k}, which—to our knowledge—represents the largest-scale 4D robot-interaction dataset. Comprising 201,426 high-fidelity episodes, Robo4D-200k provides the unprecedented data volume and interaction diversity required to train foundation models with spatiotemporal and physical awareness. More details of our dataset can be found in the supplementary material.

%% file: sec/4_experiment.tex
\section{Experiments}

\subsection{Setting}
\paragraph{Implementation Details.}
We build our framework upon the WAN 2.1 \cite{wan2025wan} base model (14B parameters), utilizing the 4D-aware pre-trained weights from 4DNex \cite{chen20254dnex}. Given the substantial scale of the backbone, we employ Low-Rank Adaptation (LoRA) \cite{hu2022lora} for parameter-efficient fine-tuning. This allows us to inject domain-specific robotic dynamics into the Diffusion Transformer \cite{dit} while preserving the powerful generative priors of the base model.
While the original WAN 2.1 includes a text encoder, we intentionally replace the text embedding with the VAE latents of robot sequences. By omitting high-level instructions, we force the network to decouple visual synthesis from semantic ambiguity and focus on the precise execution of action controls. For all other configurations, we follow the protocol established in the official 4DNex implementation.

We curated a stratified benchmark suite comprising 2\% of our Robo4D-200k dataset for the validation. More implementation details, ablations of text embeddings, and benchmark details are provided in the supplementary material.



\begin{table}[t]
\centering
\setlength{\tabcolsep}{8pt}
\caption{\textbf{Quantitative comparison of video generation metrics}. $\uparrow$ and $\downarrow$  respectively indicates higher and lower is better. \textbf{xxx} denotes the best and \underline{xxx} indicates the 2nd-best. “Action” means the representation of robot controls, where “Emb.” is token embedding. “Output” represents the modality of generation output.} 
\vspace{-0.3cm}
\label{tab:2d_metrics}
\resizebox{1.0\linewidth}{!}{
\begin{tabular}{l|cc|cccccc}
\toprule
\textbf{Method} & \textbf{Action} & \textbf{Output} & \textbf{PSNR}$\uparrow$  &  \textbf{SSIM}$\uparrow$ &
\textbf{L2$_{latent}$}$\downarrow$ &
\textbf{FID}$\downarrow$ & \textbf{FVD}$\downarrow$ &
\textbf{LPIPS}$\downarrow$ \\ \midrule
UniSim [ICLR'24] \cite{yang2024learning} & Text & RGB & 19.32 & 0.681 & 0.2120 & 32.3 & 153.2 & 0.175 
\\
IRA-Sim [ICCV'25] \cite{zhu2025irasim} & Emb. & RGB & 20.21 & 0.813 & 0.1722 & 
\underline{25.2} & 126.0 & 0.135 \\
Cosmos [arXiv'25] \cite{agarwal2025cosmos} & Emb. & RGB & 20.39 & 0.787 & 0.1935 & 27.1 & 113.4 & \underline{0.110} \\
EVAC [arXiv'25] \cite{jiang2025enerverse} & Emb.+2D & RGB & 20.88 & \underline{0.832} & 0.1896 & 29.3 & 122.0 & 0.150 \\
ORV [CVPR'26] \cite{yang2026orv} & Emb.+3D & RGB & 19.45 & 0.790 & 0.2002 & 
30.1 & 130.1 & 0.143\\
Ctrl-World [ICLR'26] \cite{guo2026ctrl} & Emb. & RGB & \underline{21.03} & 0.803 & \underline{0.1533} & \textbf{24.9} & \underline{112.8} & 0.122 \\
TesserAct [ICCV'25] \cite{zhen2025tesseract} & Text & 4D & 19.35 & 0.766 & 0.1911 & 29.5 & 120.3 & 0.158\\
\textbf{Ours} & \textbf{4D} & \textbf{4D} & \textbf{22.50} & \textbf{0.864} & \textbf{0.1380} & \underline{25.2} & \textbf{98.5} & \underline{0.105} \\ \bottomrule
\end{tabular}
}
\vspace{-0.2cm}
\end{table}

\begin{table}[t]
\centering
\setlength{\tabcolsep}{6pt}
\caption{\textbf{Quantitative comparison of geometric metrics}. “(temp)” denotes the self-temporal consistency across frames. The threshold of F-Score is set to @0.01.}
\vspace{-0.3cm}
\label{tab:3d_metrics}
\resizebox{0.9\linewidth}{!}{
\begin{tabular}{l|cc|cc|cc}
\toprule
\textbf{Method} & 
\textbf{CD-$L_1$$\downarrow$} & \textbf{CD-$L_1$ (temp)$\downarrow$} & \textbf{CD-$L_2$$\downarrow$} & \textbf{CD-$L_2$ (temp)$\downarrow$} & 
\textbf{F-Score$\uparrow$} & \textbf{F-Score (temp)$\uparrow$} \\ 
\midrule
TesserAct [ICCV'25] \cite{zhen2025tesseract} & 
0.0836 & \textbf{0.0067} &
0.0130 & 0.0008 & 
0.2896 & 0.9523 \\
\textbf{Ours} & 
\textbf{0.0479} & 0.0074 & 
\textbf{0.0077} & \textbf{0.0002} & 
\textbf{0.4733} & \textbf{0.9686} \\ 
\bottomrule
\end{tabular}
}
\vspace{-0.4cm}
\end{table}

\paragraph{Baselines.}
We benchmark our method against a diverse set of state-of-the-art generative embodied simulators with different action-conditioning strategies and output modalities. UniSim \cite{yang2024learning} pioneered the generative simulator by synthesizing visual outcomes from text instructions. IRASim \cite{zhu2025irasim}, Cosmos \cite{agarwal2025cosmos}, and Ctrl-World \cite{guo2026ctrl} utilize adaptive layers to inject robot-specific action embeddings. EVAC \cite{jiang2025enerverse} and ORV \cite{yang2026orv} introduce auxiliary cues, such as 2D angle arrows and 3D environmental occupancy grids. While most predecessors generate pure RGB video, TesserAct \cite{zhen2025tesseract} represents the current frontier by outputting 4D signals (RGB, depth, and surface normals), though it remains limited by its reliance on coarse language instructions.
As the field in its infancy, we prioritize models with open-sourced implementations to ensure a fair comparison.

\paragraph{Metrics.}
To compare with the methods that output RGB videos, we use two types of metrics: computation-based (PSNR \cite{hore2010image}, SSIM \cite{wang2004image}) and model-based (Latent L2 loss, FID \cite{heusel2017gans}, FVD \cite{ranftl2020towards} and LPIPS \cite{zhang2018unreasonable}). As both TesserAct and our method output 4D videos, we additionally measure geometric metrics, including Chamfer Distance and F-Score@0.01, considering both the accuracy with the Ground Truth and the temporal consistency across frames (indicated by “temp”).

\begin{figure}[t]
\centering
\includegraphics[width=1.0\linewidth]{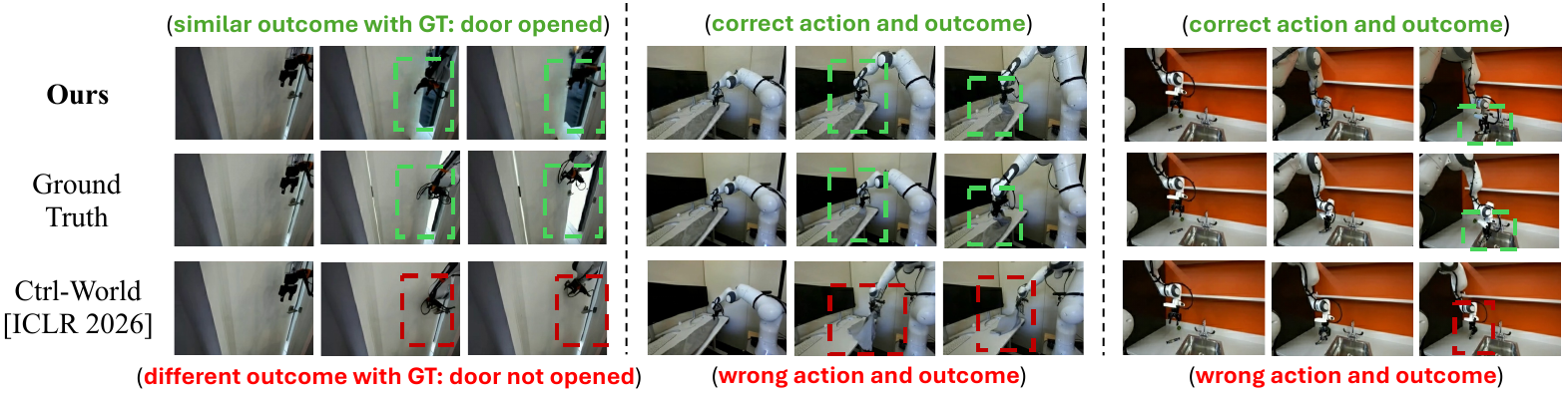}
\vspace{-0.4cm}
\caption{\textbf{Qualitative comparison of 2D video synthesis} between our Kinema4D and Ctrl-World \cite{guo2026ctrl}. Our method achieves superior fidelity. In contrast, Ctrl-World exhibits distorted actions and unrealistic environmental transitions.}
\label{fig:qualitative_result_2D}
\vspace{-0.1cm}
\end{figure}

\begin{figure}[!h]
\centering
\includegraphics[width=0.7\linewidth]{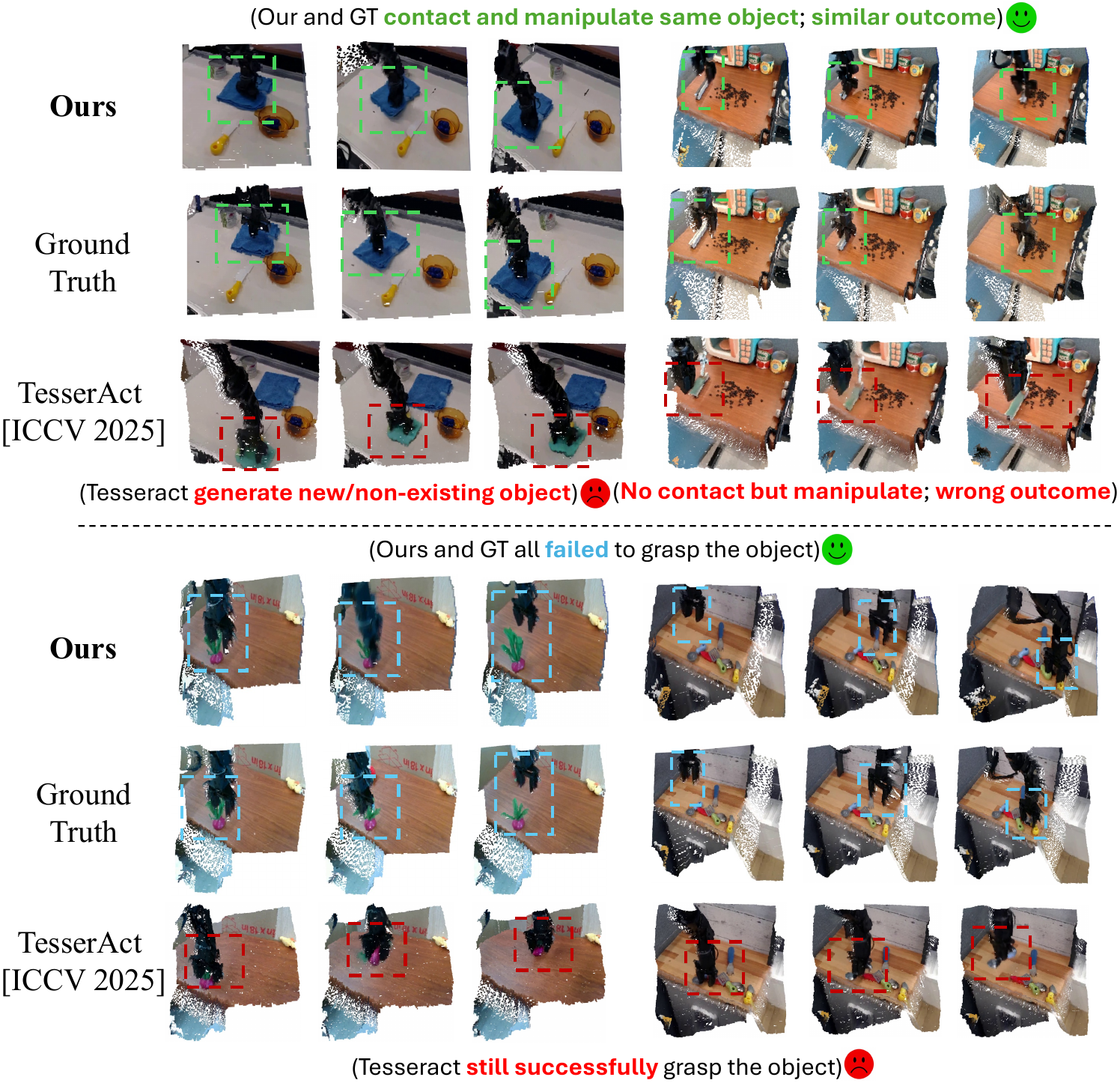}
\vspace{-0.2cm}
\caption{\textbf{4D qualitative comparison between our Kinema4D and TesserAct \cite{zhen2025tesseract}}. Unlike TesserAct that hallucinate outcomes, our Kinema4D precisely reflects Ground-Truth executions, including “near-miss” failure cases. For example, in the bottom left corner example, our model \textit{correctly interprets the spatial gap} between the gripper and the plant, \textit{even when their \textbf{RGB textures overlap in 2D views}}.}
\label{fig:qualitative_result_4D}
\vspace{-0.5cm}
\end{figure}

\subsection{Main Results}

\paragraph{Quantitative results.}
The quantitative evaluation of video synthesis quality is summarized in \cref{tab:2d_metrics}. Our Kinema4D is uniquely characterized by its ability to precisely represent actions and synthesize outcomes within a unified 4D space. This advantage allows our framework to achieve either the leading or second-best performance across all metrics. The geometric fidelity of the generated 4D sequences is detailed in \cref{tab:3d_metrics}. While TesserAct \cite{zhen2025tesseract} exhibits slightly better performance in temporal CD-$L_1$, our approach outperforms TesserAct across all other metrics. When evaluated against absolute ground-truth rather than mere self-consistency, our method shows a significant margin of superiority.

\paragraph{Qualitative results.}
We compare our method with the best baseline that output 2D videos (\ie, Ctrl-World \cite{guo2026ctrl}), focusing on the qualitative results under 2D video modality. As shown in \cref{fig:qualitative_result_2D}, our Kinema4D demonstrates a superior ability to synthesize high-fidelity sequences that strictly adhere to input action trajectories while generating physically consistent environmental responses that are highly similar to the ground truth. In contrast, as Ctrl-World is limited to 2D video synthesis and relies on implicit robot embeddings, it often produces distorted robot kinematics and unrealistic environmental changes.

In \cref{fig:qualitative_result_4D}, we present a 4D qualitative comparison between our framework and TesserAct \cite{zhen2025tesseract}. As TesserAct relies on text instructions, it struggles to align with the actual robot actions, and often yields incorrect environmental outcomes.
Instead, by utilizing precise 4D controls, our Kinema4D demonstrates the ability to simulate \textit{\textbf{both successful} rollouts and \textbf{subtle execution failures}}. \textit{\textbf{Most notably}}, even when the RGB textures of the gripper and the object appear to “contact” from a 2D perspective, our 4D-aware framework correctly interprets the spatial gap and simulates a `near-miss' or a failed grasp.
More animated results are presented in the supplementary material.

\subsection{Policy Evaluation}
We investigate the utility of our Kinema4D as a high-fidelity tool for evaluating robotic policies—if the simulator can accurately simulating realistic outcomes of executing a policy's rollout.
We deploy our evaluation across both standardized simulation platforms (noise-free) and real-world (complex physics) environments.

\paragraph{Setup and implementations.}
In the simulation platform, we execute a Diffusion Policy \cite{chi2025diffusion} within the LIBERO \cite{libero} to generate a sequence of action rollouts, then synthesize the environmental outcomes using our Kinema4D. Since the robot mesh and camera rendering are integrated within the platform, we can easily get the robot pointmap without any reconstruction or calibration, showing the simplicity of deploying our Kinematic Control in the simulation platform.
By leveraging native simulation parameters, this evaluation protocol isolates external noise and establishes a standard setup, ensuring that the benchmarking of future methods is conducted under a strictly equitable framework.

As for real-world evaluation, we deploy a physical robot setup in a laboratory environment to evaluate the zero-shot generalization of our framework. We execute a Diffusion Policy \cite{chi2025diffusion} on the hardware to provide rollouts. Next, we follow the reconstruction pipeline described in \cref{sec:kinematics_control} to get the robot pointmap and then synthesize the resulting world transitions. Notably, our experiments are conducted \textit{\textbf{without} any fine-tuning on real-world data}; the physical environment remains \textit{entirely \textbf{out-of-distribution} (OOD)}. Unlike existing simulators rely on real-world finetuning \cite{zhu2025irasim}, or use the real-world setup same with training data \cite{guo2026ctrl}, we represent the \textbf{\textit{first-time}} evaluation of a embodied world model under rigorous OOD conditions. We omit qualitative comparisons with baselines here, as their performance degrades significantly.
For both the simulation and real-world evaluation, we set 3 different evaluation scenarios with 50 runs each.

\paragraph{Results.}
The quantitative success rates are summarized in \cref{tab:policy}. Within the simulation platform, our framework achieves a success rate closely aligned with the ground truth. Since our training data includes LIBERO-produced data, this proves that our method effectively captures the in-domain knowledge. In the real-world (OOD) evaluation, while we observe \textit{a comparatively wider gap} between our simulated and actual results, the discrepancy \textit{remains within a reasonable margin}. This suggests the challenges of zero-shot transfer. Moreover, the success rates synthesized by our model are higher than actual executions, highlighting that simulating complex failure modes remains more challenging. 

\begin{figure}[t]
\centering
\includegraphics[width=1.0\linewidth]{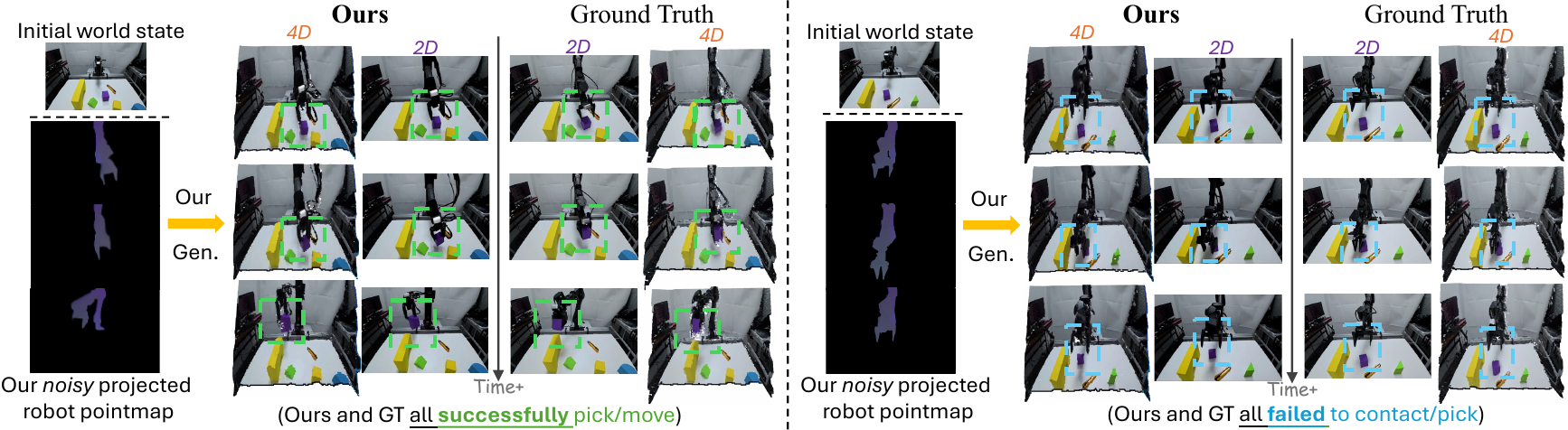}
\vspace{-0.3cm}
\caption{\textbf{Qualitative comparison between real-world Ground Truth (GT) and our method for zero-shot OOD evaluation}. To intuitively compare with GT, we also reconstruct the GT using \cite{xiao2025spatialtracker} and present both 4D and 2D results. Our results strongly align with real outcomes, accurately synthesizing both successful rollouts and “near-miss” failures. For instance, in the right figure, our model \textit{correctly interprets the spatial gap} between the gripper and the cube, \textit{even when their \textbf{RGB textures overlap in 2D views}}.
Besides, the \textit{noisy} robot pointmap produced by our reconstruction pipeline is shown, proving the robustness of our framework.}
\label{fig:qualitative_result_real}
\vspace{-0.4cm}
\end{figure}

The qualitative performance in real-world environments is illustrated in \cref{fig:qualitative_result_real}, highlighting our two critical strengths:
\textbf{i)} Consistent with \cref{fig:qualitative_result_4D}, even in scenarios where the 2D RGB textures of the gripper and object overlap, our Kinema4D correctly interprets the spatial gap and synthesizes `near-miss' failures.
\textbf{ii)} We also visualize the robot pointmaps produced by our real-world reconstruction pipeline (\cref{sec:kinematics_control}). Despite the presence of noisy artifacts, our model refines these irregularities to produce coherent and physically plausible videos. This shows the robustness and real-world deployability of our kinematic-controlled framework.
Due to the space limit, the qualitative results of simulation and more details of our evaluation are presented in the supplementary material.

\begin{table}[t]
\centering
\setlength{\tabcolsep}{6pt}
\caption{Policy evaluation of actual pick\&place under 3 different setups of both simulation platform and real-world (zero-shot OOD) environment. “Ground Truth” denotes the actual execution, “Ours” indicates our simulation, and “Diff” is their difference.}
\vspace{-0.3cm}
\label{tab:policy}
\resizebox{0.5\linewidth}{!}{
\begin{tabular}{l|ccc|ccc}
\toprule
\textbf{Evaluator} & \multicolumn{3}{|c|}{Simulation} & \multicolumn{3}{|c}{Real-world (OOD)}\\
& 1 & 2 & 3 & 1 & 2 & 3 \\
\midrule
Ground Truth & 
0.48 & 0.38 & 0.80 
& 0.34 & 0.46 & 0.78 \\
Ours &
0.56 & 0.46 & 0.84 
& 0.60 & 0.76 & 0.90 \\
\hline
Diff &
0.08 & 0.08 & 0.04 
& 0.26 & 0.30 & 0.12 \\ 
\bottomrule
\end{tabular}
}
\end{table}

\subsection{Ablation Studies and Analysis}
We provide comprehensive ablation analysis and report results in \cref{tab:ablation}.

\paragraph{Different representations of robot controls.}
We replace our robot pointmap conditions with other types of robot control representations, including text instructions (“text”), robot binary masks (“binary”), token embeddings (“emb.”), robot RGB (“RGB”) and robot RGB+pointmap (“RGB+pm”) sequence. Our pointmap representation yields the 2nd-best results. Interestingly, using RGB+pointmap sequence yields marginal improvements, as RGB data potentially introduces more noise and overfitting risks. Instead, pointmap already captures the essential spatial information required for precise control.

\paragraph{Embodiment-agnostic modeling enables scalable training.}
Unlike raw action vectors that are sensitive to embodiment morphology, our pointmap-based control enables embodiment-agnostic modeling by decoupling kinematic intent from specific hardware configurations. To validate this, we compare our mixed-dataset training against a single-domain baseline trained and tested exclusively on Droid \cite{khazatsky2024droid}. The baseline (“single”) exhibited degraded performance. This confirms that our approach effectively leverages data diversity to enhance generalization, demonstrating a significant \textit{scaling advantage} over domain-specific modeling.

\paragraph{Why not generate RGB videos first, then reconstruct?}
We trained our method to purely output RGB videos (no pointmap, “2D-out”) and then reconstruct the produced RGB videos using ST-v2 \cite{xiao2025spatialtracker}. This significantly degrades the performance, proving the necessity of 4D awareness throughout the generative process.

\paragraph{The robot mask map.}
As in \cref{sec:4d_gen_modeling}, we introduced a robot occupancy mask and softly set 10\% occupied regions to 0.5.
Here we trained our method without robot mask or using different soft mask ratio. Our method performs stably well under different ratios, yet discarding the mask or setting 0\% regions to 0.5 degrades the performance, showing the necessity of generative refinement.

\paragraph{The robustness to the robot pointmap noise.}
We disturb the produced robot pointmap by randomly removing (5\%), adding gaussian noise (“gaus.”), translating ($\pm5$ pixel coordinates, “trans.”) and rotating ($\pm5^\circ$, “rot.”). Our framework is robust to the noise of robot pointmap control, potentially due to the prior of the initial world image and the refinement of generative modeling.

\begin{table}[t]
\centering
\setlength{\tabcolsep}{3pt}
\caption{Quantitative results of ablation studies.}
\vspace{-0.3cm}
\label{tab:ablation}
\resizebox{1.0\linewidth}{!}{
\begin{tabular}{l|c|ccccc|c|c|ccccc|cccc}
\toprule
Metrics & 
Ours & 
text & binary & emb. & RGB & RGB+pm &
single & 
2D-out &
w/o mask & 0\% & 20\% & 50\% & 70\% &
remove & gaus. & trans. & rot.\\ 
\midrule
\textbf{PSNR}$\uparrow$ & 
22.50 & 
19.89 & 21.47 & 20.89 & 21.53 & \textbf{22.98} &
21.26 & 
20.07 &
21.03 & 21.10 & 22.04 & 21.75 & 21.83 &
22.48 & 21.98 & 21.87 & 22.34\\ 
\textbf{FID$\downarrow$}  & 
25.2 & 
28.8 & 27.5 & 26.3 & 25.8 & 25.7 &
26.7 & 
27.4 &
26.8 & 26.1 & 25.2 & \textbf{25.0} & 26.0 &
26.0 & 25.3 & 25.6 & 26.3\\
\textbf{CD-$L_1$$\downarrow$}  & 
0.0479 & 
0.0750 & 0.0639 & 0.0528 & 0.0677 & 0.0495 &
0.0581 & 
0.0712 &
0.0510 & 0.0528 & \textbf{0.0433} & 0.0455 & 0.0463 &
0.0499 & 0.0501 & 0.0513 & 0.0483\\
\bottomrule
\end{tabular}
}
\vspace{-0.4cm}
\end{table}

%% file: sec/5_conclusion.tex
\section{Conclusion}
We presented Kinema4D, a novel framework that shifts the paradigm of robotic simulation to 4D spatial-temporal reasoning. By integrating a kinematics-driven grounding stage with a diffusion transformer-based generative pipeline, we effectively decouple deterministic robot motion from stochastic environmental reactions. Experimental results demonstrate that Kinema4D can generalize to simulate diverse real-world dynamics. By providing a 4D world, Kinema4D paves the way for scalable, high-fidelity, and complex embodied simulations.

\paragraph{Limitations.} Since the environmental dynamics are learned through statistical synthesis rather than governed by explicit physical constraints (such as rigid-body dynamics or friction coefficients), our method may occasionally produce behaviors that violate conservation laws or exhibit penetration artifacts, leaving a future research direction of incorporating physical laws into our model.

%% file: sec/x_supple.tex
\clearpage
\begin{appendices}
\section*{\Large \textit{Supplementary Material} for Kinema4D}
\renewcommand{\thesection}{\Alph{section}}%
\renewcommand\thetable{\Roman{table}}
\renewcommand\thefigure{\Roman{figure}}

\centerline{\large{\textbf{Outline}}}
\noindent{This supplementary document is arranged as follows:}\\
(1) Sec.~\ref{sec:more_qualitative_results} provides more qualitative results;\\
(2) Sec.~\ref{sec:more_details} elaborates more technical details and discussions of our framework implementation, dataset, and real-world deployment;\\
(3) Sec.~\ref{sec:more_ablations} discusses more ablation studies and analysis.\\

\section{More Qualitative Results}
\label{sec:more_qualitative_results}

\begin{figure}[t]
\centering
\includegraphics[width=1.0\linewidth]{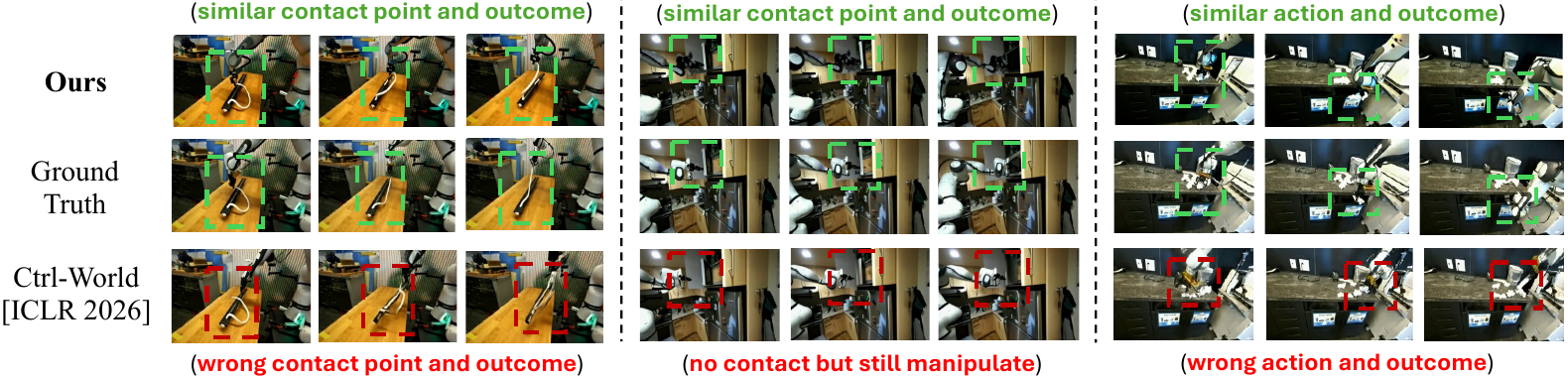}
\caption{\textbf{Qualitative comparison of 2D video synthesis} between our Kinema4D and Ctrl-World \cite{guo2026ctrl}. Our method achieves superior fidelity. In contrast, Ctrl-World exhibits distorted actions and unrealistic environmental transitions.}
\label{fig:qualitative_result_2D_supple}
\end{figure}

\begin{figure}[!ht]
\centering
\includegraphics[width=0.8\linewidth]{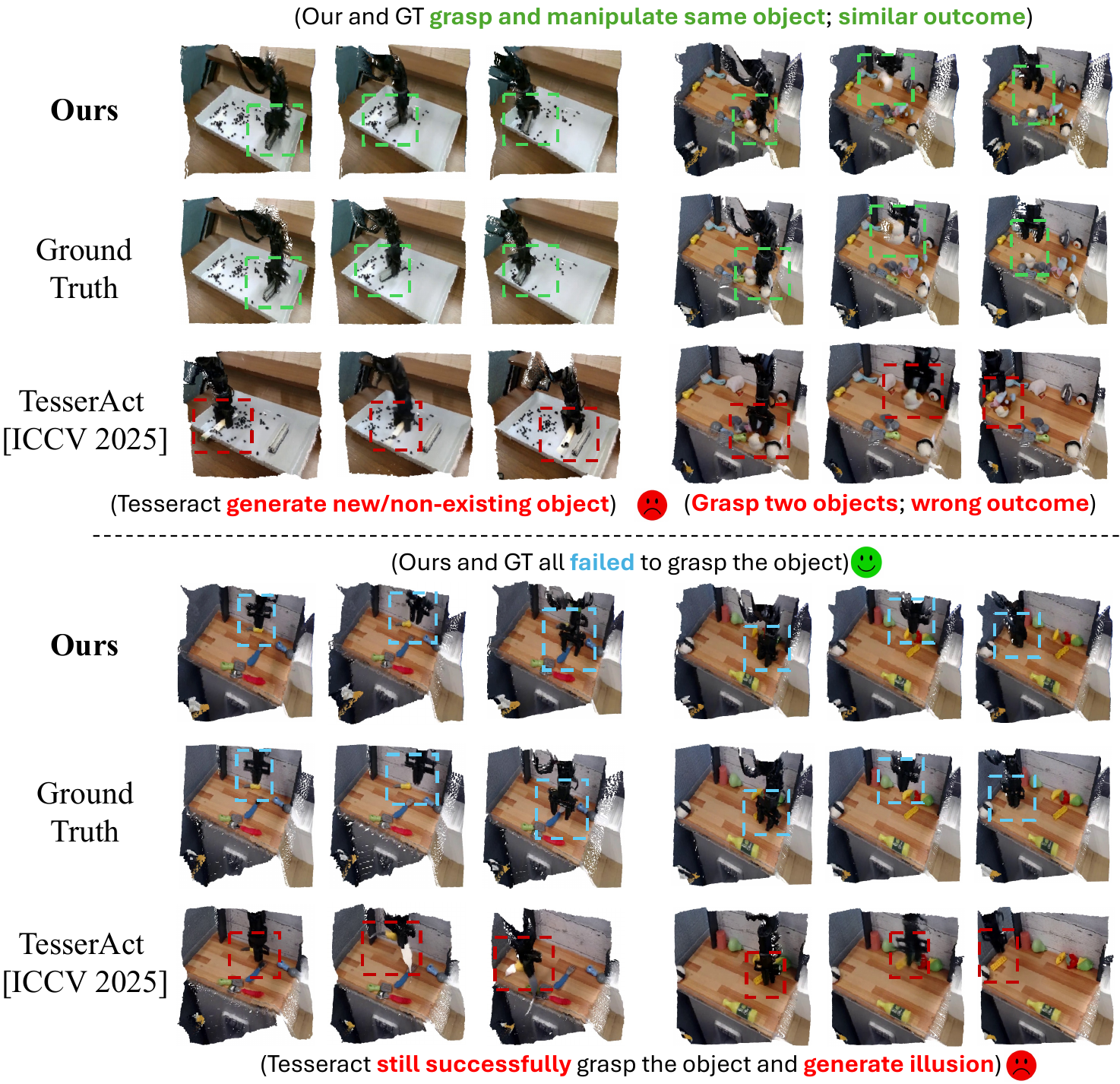}
\caption{\textbf{4D qualitative comparison between our Kinema4D and TesserAct \cite{zhen2025tesseract}}. Unlike TesserAct that hallucinate outcomes, our Kinema4D precisely reflects Ground-Truth executions, including “near-miss” failure cases. For example, in the two bottom examples, our model \textit{correctly interprets the spatial gap} between the gripper and the objects, \textit{even when their \textbf{RGB textures overlap in 2D views}}.}
\label{fig:qualitative_result_4D_supple}
\end{figure}

\begin{figure}[t]
\centering
\includegraphics[width=1.0\linewidth]{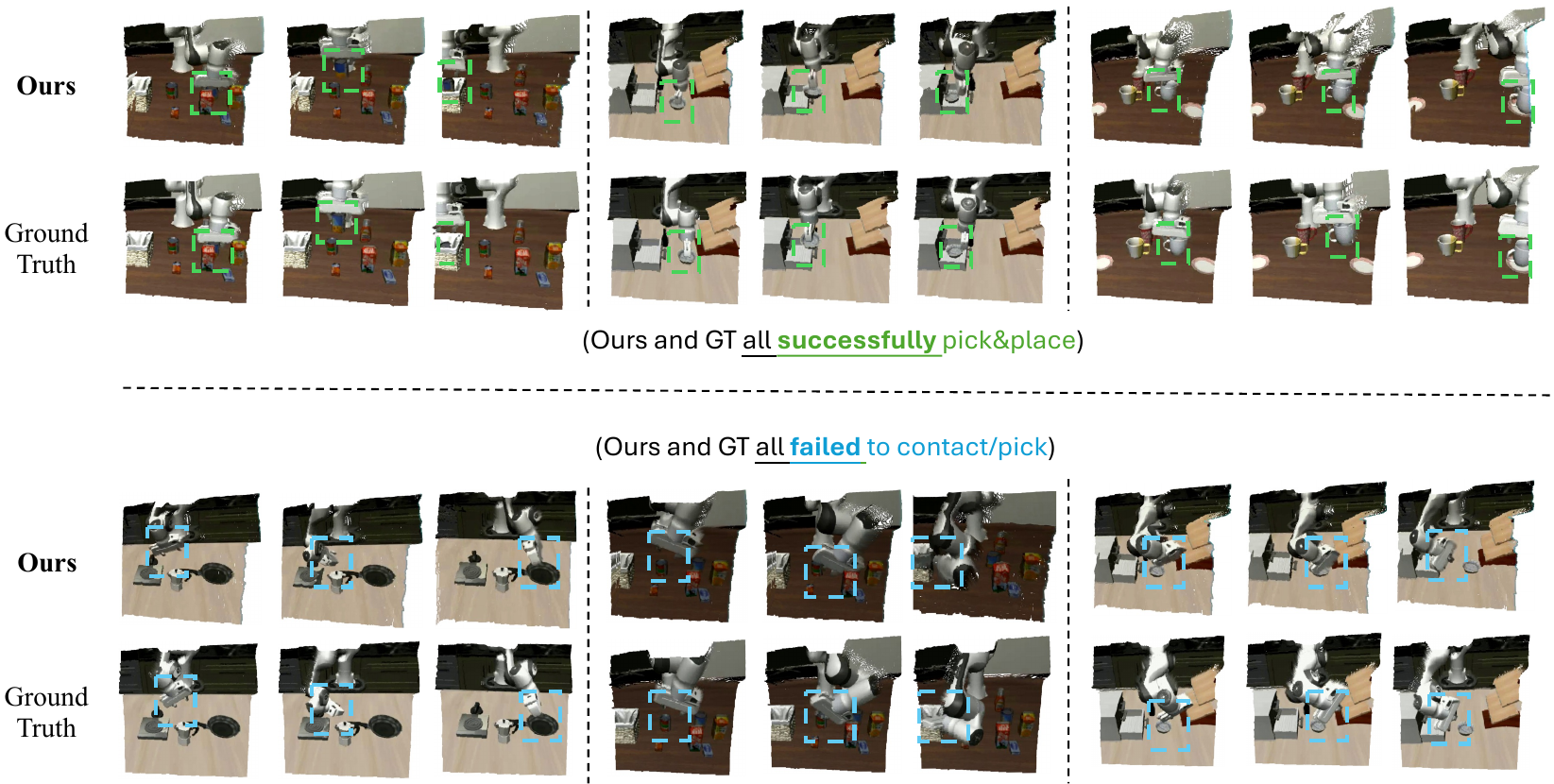}
\caption{\textbf{4D qualitative results of our Kinema4D for policy evaluation in simulation platform}. Our simulated results strongly align with actual executions, accurately synthesizing both successful rollouts and “near-miss” failures. For instance, in the bottom figures, our model correctly interprets the spatial gap between the gripper and the object, even when their RGB textures overlap in 2D views.}
\label{fig:qualitative_result_sim}
\end{figure}

\begin{figure}[t]
\centering
\includegraphics[width=1.0\linewidth]{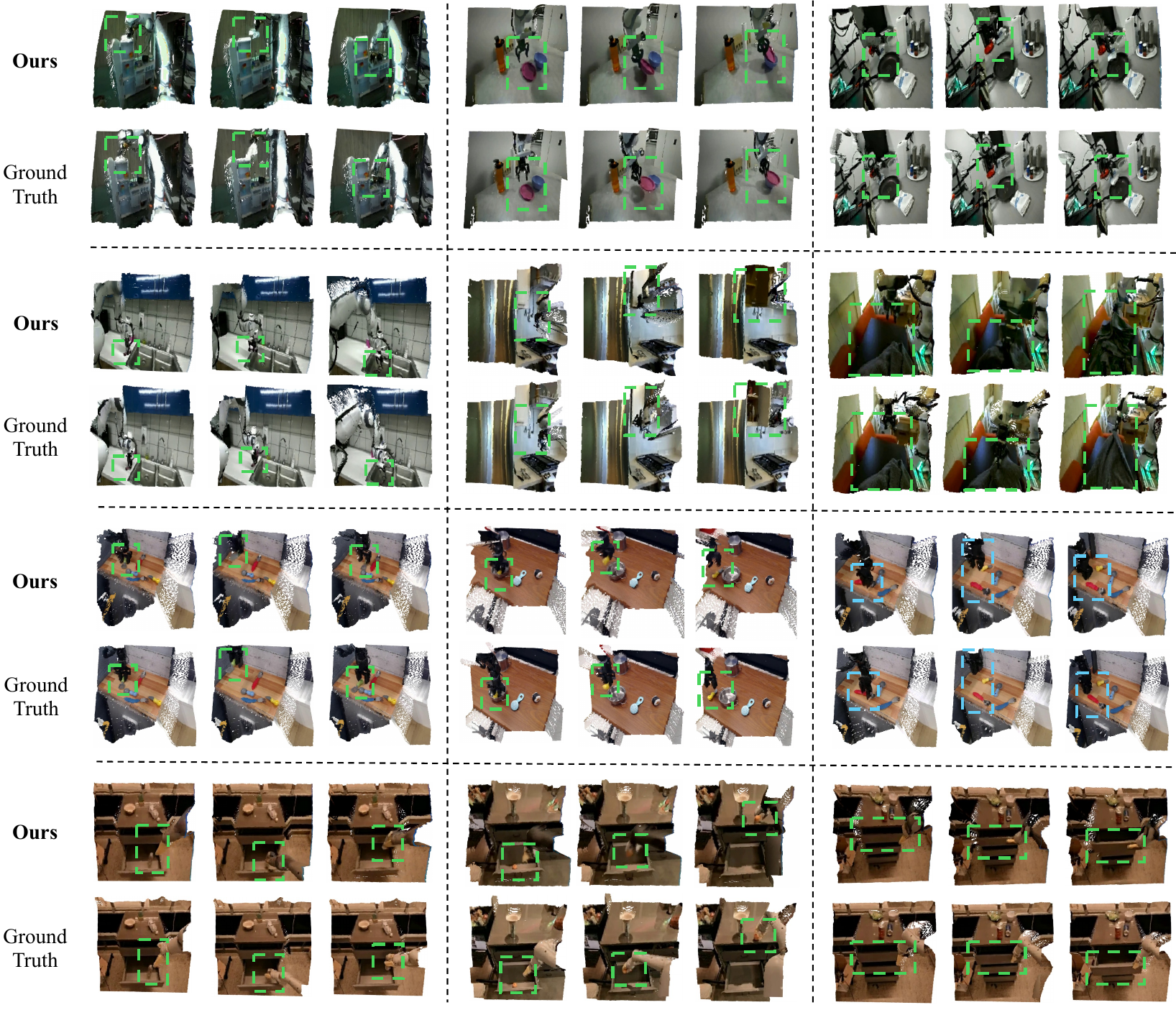}
\caption{\textbf{Extensive 4D qualitative results of our Kinema4D}, across diverse domains, environments, robot actions, and the manipulated objects. These results are \textit{non-overlapping} with other figures. More animated results are provided in \textit{demo.mp4}.}
\label{fig:qualitative_result_all}
\end{figure}

\cref{fig:qualitative_result_2D_supple} shows more 2D qualitative comparison with Ctrl-World \cite{guo2026ctrl}, \cref{fig:qualitative_result_4D_supple} visualizes more 4D qualitative comparison with TesserAct \cite{zhen2025tesseract}, \cref{fig:qualitative_result_sim} illustrates our qualitative results for policy evaluation in simulation platform, and \cref{fig:qualitative_result_all} provides extensive qualitative results of our method with Ground Truth as the reference. These results again demonstrate the effectiveness of our Kinema4D.

We also provide a demo video, \textbf{\textit{demo.mp4}}, for our work. In this video, a teaser introduction is provided from the beginning to 1:20. 1:20 $\sim$ 2:40 illustrates our framework, 2:40 $\sim$ 3:46 describes our dataset acquisition, 3:50 $\sim$ 7:00 provides quantitative and animated qualitative results for the main comparison, 7:00 $\sim$ 8:18 shows animated qualitative results for policy evaluation, 8:20 $\sim$ 8:30 visualizes extensive animated results.

\section{More Technical Details}
\label{sec:more_details}

\subsection{Framework Implementations.}
\paragraph{Deployment of Kinematic Control.} 
In the initial Kinematics Control phase, we implement a multi-view reconstruction-projection pipeline. For individual robot reconstruction, we capture sparse orbit videos via a smartphone and sample 20 representative frames. We employ Grounded-SAM2$^{\ref{footnote:groundedsam_repo}}$ \cite{liu2023grounding, ravi2024sam2} to segment the target object in the initial frame using the prompt \texttt{“the object on the desk”}, a strategy that proved highly robust across varying lighting conditions. 
\footnotetext[1]{\label{footnote:groundedsam_repo}\url{https://github.com/IDEA-Research/Grounded-SAM-2}}
These segments are then propagated throughout the sequence using the temporal tracking capabilities of SAM2$^{\ref{footnote:sam2_repo}}$ \cite{ravi2024sam2}.
\footnotetext[2]{\label{footnote:sam2_repo}\url{https://github.com/facebookresearch/sam2}}
Given these masked multi-view images, we utilize ReconViaGen$^{\ref{footnote:reconviagen_repo}}$ \cite{chang2025reconviagen} for pose-free high-fidelity shape synthesis. 
\footnotetext[3]{\label{footnote:reconviagen_repo}\url{https://github.com/estheryang11/ReconViaGen}}
Compared to traditional Structure-from-Motion (SfM) pipelines like COLMAP \cite{schonberger2016structure,schonberger2016pixelwise} or recent feed-forward models like VGGT \cite{wang2025vggt}, ReconViaGen generates more complete meshes with significantly fewer holes in unobserved regions—a critical advantage under sparse, unconstrained viewpoints. Furthermore, it outperforms commercial-grade alternatives such as Hunyuan3D \cite{yang2024hunyuan3d} in our specific laboratory setting. This pipeline produces a high-quality textured mesh and precise camera poses in approximately \textit{\textbf{15} seconds} on a single NVIDIA A100 GPU, facilitating the rapid construction of robot pointmap sequences in real-world deployment. 

Throughout both training and inference, we normalize each robot pointmap to a $[0, 1]$ range based on the extrema (minimum and maximum values) across the entire pointmap sequence prior to VAE encoding. This sequence-level normalization facilitates affine-invariant 4D control, encouraging the framework to prioritize relative spatial geometry and inter-object relationships over absolute coordinate values. By mapping diverse robot trajectories into a canonical latent space, this approach enhances the model's ability to generalize across different scales and workspace configurations.

\paragraph{Details of 4D Generative Modeling.} 
In the second phase of 4D Generative Modeling, we employ Parameter-Efficient Fine-Tuning (PEFT) via LoRA \cite{hu2022lora}, initializing weights from 4DNeX$^{\ref{footnote:4dnex_repo}}$ \cite{chen20254dnex}.
\footnotetext[4]{\label{footnote:4dnex_repo}\url{https://github.com/3DTopia/4DNeX}}
The base model, 4DNeX, was originally fine-tuned on the Wan2.1 \cite{wan2025wan} 14B image-to-video architecture. To mitigate the distribution shift between RGB and pointmap latents, we adopt the latent normalization strategy from 4DNeX. Specifically, we compute the empirical statistics of the pointmap domain in the latent space over 5,000 random training samples, yielding a mean $\mu = -0.17$ and a standard deviation $\sigma = 1.36$. These values serve as constant normalization terms for the XYZ latents during both training and inference. The LoRA fine-tuning is configured with a rank of 64 to balance parameter efficiency with representational capacity. Training is conducted with a batch size of 32 using the AdamW optimizer, a peak learning rate of $2 \times 10^{-5}$, and a cosine warmup followed by constant decay. The model is trained for 5,000 iterations across 32 NVIDIA A100 GPUs, taking approximately 2 days, at a spatial resolution of $480 \times 720$ per modality, occupying 67GB of memory per GPU. While the original 4DNeX or WAN uses a text encoder to inject text conditions, we intentionally replace the text embedding with the VAE latents of robot sequences, which is discussed and ablated in \cref{sec:more_ablations} (“Text conditions”).
During training and inference, our robot mask map is spatially and temporally downsampled according to the VAE latent transformation rules, ensuring exact alignment with the latent tensors. It is subsequently expanded to four channels (yielding a shape of $B \times 4 \times H_{latent} \times W_{latent}$), where non-zero values are restricted to robot-occupied positions, and is then channel-wise concatenated with both the noisy and input latents. In addition, we follow ControlNet \cite{zhang2023adding} to add a 1 × 1 zero-initialized convolution layer on the robot latent, to eliminate random noise as gradients in the initial finetuning steps.

\subsection{Dataset}
\paragraph{Acquisition of LIBERO simulated data.}
Our data generation framework is built upon the MuJoCo \cite{todorov2012mujoco} engine within the LIBERO \cite{libero} platform. We extend the OpenVLA \cite{kim24openvla} dataset pipeline$^{\ref{footnote:openvla_repo}}$ by integrating MuJoCo’s depth-buffer rendering to capture per-pixel depth maps alongside RGB sequences. 
\footnotetext[5]{\label{footnote:openvla_repo}\url{https://github.com/openvla/openvla}}
These are subsequently back-projected into world-coordinate pointmaps using intrinsic and extrinsic camera parameters. To maintain a consistent data format, these pointmaps are stored as pseudo-RGB MP4 videos after undergoing per-sequence min-max normalization. To ensure high data variance, we perform procedural randomization of background objects within each scene. While the core dataset consists of successful demonstrations, we systematically synthesize failure cases to enhance the model's discriminative capabilities. Specifically, each successful trajectory is partitioned into three temporal segments, where Gaussian noise of varying intensities ($\sigma \in \{0.5, 0.8, 1.2\}$) is injected into the 6-DoF pose dimensions. This process yields nine distinct failure trajectories per demonstration, as the cumulative error causes the robot to deviate progressively from the target path. Notably, the gripper state remains unperturbed to ensure that the resulting failures stem from reaching or positioning errors rather than unnatural grasping artifacts.

\paragraph{Extracting robot pointmap from 4D data annotations.}
Using either simulation platform for producing synthetic data or ST-v2 \cite{xiao2025spatialtracker} for processing real-world data, we can easily gain the whole pointmap (including both the robot and environment) sequence at a main camera view. To further gain the robot-only pointmap as the control signal, we need to extract the robot area from the given pointmap. To realize this, we segment the robot’s presence within RGB frames by utilizing SAM2 \cite{ravi2024sam2} with a robust semantic prompt (\eg, “the robot arm”), and apply the segmented mask on both pointmap and RGB frames to gain robot-only sequences. 
By doing so, we maintain the automated property of our data processing pipeline, so that our framework can leverage scalable video data from diverse, in-the-wild sources without manual labeling overhead. During both the model training and inference, we use the processed robot pointmap sequence as the control signal.

\paragraph{Language annotations.}
As we find that many text instruction labels are incomplete in our original data resource \cite{khazatsky2024droid,walke2023bridgedata,brohan2022rt}. We utilize Qwen3-VL-plus$^{\ref{footnote:qwen3_repo}}$ \cite{Qwen3-VL} to generate the language descriptions for each episode. 
\footnotetext[6]{\label{footnote:qwen3_repo}\url{https://github.com/QwenLM/Qwen3-VL}}
Two types of texts are generated, including the complete text (considering the whole robot-world dynamic interactions across the whole video) and the environment-centric descriptions (only considering the static environment at the initial frame).

\paragraph{Validation split.}
To ensure a rigorous and fair comparison, we constructed a unified validation set by sampling and aggregating 3200 episodes from the official validation episodes of all baselines. This evaluation set is sampled from our core data sources—2,000 from DROID, 1,000 from Bridge, 100 from RT-1, and 100 from LIBERO (50 successful and 50 failure modes)—preserving the proportional distribution of the training set. This composition ensures comprehensive coverage of complex robot-world interactions, ranging from articulated/deformable object manipulation and pick-and-place of objects. Crucially, all samples within this combined validation set were strictly excluded from our training set to prevent data leakage. This setup allows us to evaluate the generalization capability of our model across multiple data distributions simultaneously, whereas baselines are often specialized to a single domain. 

\paragraph{The underlying logic behind 4D pseudo annotation.}
To facilitate efficient large-scale data synthesis, we employ ST-v2 \cite{xiao2025spatialtracker} to generate pseudo-annotations directly from diverse robotic demonstration datasets. Our core design philosophy \textit{prioritizes \textbf{scalability}} as the primary driver for building a \textit{generalizable} robotic simulator. While we acknowledge that 4D reconstruction methods like ST-v2 may not achieve absolute, sub-millimeter geometric ground truth, they provide sufficiently high-fidelity relative spatial geometry. This level of accuracy is adequate for our generative framework to internalize the underlying spatial constraints and physical affordances of the environment (as verified in our qualitative results). By prioritizing the breadth of data over the exhaustive precision of individual samples, we enable the model to learn robust, generalizable motion priors that transcend the noise inherent in pseudo-labels.

\paragraph{Frame number}.
Each raw video is sliced into 49 frames per sample and maintains a unified frequency of motion dynamics for stable training.

\subsection{Real-World Deployment}
\paragraph{Reconstruction and projection.}
Our real-world evaluation platform utilizes a YAM Arm$^{\ref{footnote:yam_arm}}$ (6-DoF collaborative manipulator). 
\footnotetext[7]{\label{footnote:yam_arm}\url{https://i2rt.com/collections/yam-arm}}
For comprehensive rollout recording and visual input, a primary RGB camera $cam$ is positioned at an oblique top-down angle relative to the workspace $\mathcal{W}$, ensuring a clear view of the manipulation area. After capturing orbit videos of the workspace, the primary viewpoint (fixed camera) image is integrated with the orbit-scan frames and processed via ReconViaGen \cite{chang2025reconviagen} to produce 3D robot mesh and the camera pose of the primary viewpoint. To ensure the robot's digital twin accurately reflects the physical state, we perform a rigid-body transformation $\mathbf{T}_{recon}^{base}$ between the physical robot's mounting point at $\mathcal{F}_{base}$ and the reconstructed global mesh at $\mathcal{F}_{recon}$ via manual calibration. 
By retrieving the real-time joint encoders from the YAM Arm’s URDF, we solve the Forward Kinematics (FK) to update the pose of each link mesh $\mathbf{x}_{base}$ at $\mathcal{F}_{base}$ and transform it into $\mathcal{F}_{recon}$: $\mathbf{x}_{recon} = \mathbf{T}_{recon}^{base} \cdot \mathbf{x}_{base} $. 

\paragraph{Experimental setup.}
We designed three distinct pick-and-place scenarios, denoted as $\mathcal{S} \in \{ \text{Easy, Medium, Hard} \}$, to evaluate the zero-shot generalization of our Kinema4D. Each scenario consists of $N=50$ trials to ensure statistical significance. The task complexity is categorized by the density of distractor objects $\mathcal{D}$ and the spatial constraints $\mathcal{C}$ within the workspace $\mathcal{W}$. Critically, both the physical embodiment and the unseen laboratory environment are entirely out-of-distribution (OOD) relative to our training data. This rigorous setup evaluates the framework's ability to transcend the reality gap without any real-world fine-tuning or domain-specific adaptation.

\begin{table}[htbp]
\centering
\setlength{\tabcolsep}{8pt}
\caption{Specifications of Real-World Evaluation Scenarios.}
\label{tab:real_world_scenarios}
\resizebox{0.8\linewidth}{!}{
\begin{tabular}{@{}lccc@{}}
\toprule
\textbf{Scenario} & \textbf{Difficulty} & \textbf{Distractors } ($\mathcal{D}$) & \textbf{Constraints / Challenges} \\ \midrule
I. Canonical & Easy & $\emptyset$ & Basic OOD texture grasping \\
II. Cluttered & Medium & $n \in [3, 5], r \leq 5\text{cm}$ & Spatial proximity \& collision avoidance \\
III. Dynamic & Hard & $n > 5$, randomized & Partial occlusion \& narrow passages \\ \bottomrule
\end{tabular}
}
\end{table}

\section{More Ablation Studies and Analysis}
\label{sec:more_ablations}

\begin{table}[t]
\centering
\setlength{\tabcolsep}{8pt}
\caption{Quantitative results of more ablation studies.}
\label{tab:more_ablation}
\resizebox{0.6\linewidth}{!}{
\begin{tabular}{l|c|c|cc|c}
\toprule
Metrics & 
Ours & 
add env. pm & 
env. text & null text & 
depth \\ 
\midrule
\textbf{PSNR$\uparrow$} & 
\textbf{22.50} & 
21.93 & 
20.97 & 22.31 & 
21.04 \\ 
\textbf{FID$\downarrow$}  & 
25.9 & 
26.2 & 
27.0 & \textbf{25.6} & 
26.3 \\
\textbf{CD-$L_1$$\downarrow$}  & 
\textbf{0.0479} & 
0.0420 & 
0.0592 & 0.0501 & 
0.0583 \\
\bottomrule
\end{tabular}
}
\end{table}

\paragraph{Why not also use the initial environmental pointmap as additional input?}
During the Kinematics Control phase, our primary objective is the extraction of precise 4D robotic trajectories. Consequently, we prioritize reconstructing the robot's geometry from masked multi-view images rather than the entire scene. This produces robot-centric pointmap sequences, which are fused with the initial world RGB frame as the conditioning input for generative modeling. While it is technically feasible to reconstruct the full static environment and provide an initial-frame world pointmap for enhanced 4D spatial grounding, our empirical results suggest this is unnecessary.
As listed in \cref{tab:more_ablation} (“add env. pm”), incorporating the complete initial-frame pointmap does not yield performance gains. We attribute this to the pre-trained weights from 4DNeX \cite{chen20254dnex}, which already endow our framework with sufficient spatial inference capabilities derived from RGB signals alone.

To further validate this design, we extended this ablation to our real-world setup. By reconstructing the entire physical environment, we provided the model with an initial real-world environmental pointmap. However, this led to a significant performance degradation, with a larger discrepancy between the policy simulation and the ground truth. Specifically, it yields [0.70, 0.82, 0.96] success rates for three different real-world test setups, respectively.
We hypothesize that full-scene reconstruction introduces aleatoric noise and sensor artifacts from the physical world, which compromises the robustness of the latent control signal. This ablation confirms that our original design—focusing on precise robot kinematics while relying on RGB for environmental context—optimizes the balance between spatial accuracy and system robustness.

\paragraph{Text conditions.}
Since our model is designed to simulate the deterministic physical outcomes of actual robotic actions that could be either successful or failed for a task execution, relying solely on high-level task instruction texts is infeasible. To investigate the impact of language conditioning, we evaluated three variants: Environment-centric text, which describes only the static scene geometry; Null conditioning (“null text”); and our proposed robot VAE latents.

As shown in \cref{tab:more_ablation}, we observed that incorporating environment-centric text (“env. text”) slightly degrades performance, likely due to the modality gap between abstract linguistic descriptions and the requirement for precise 4D transitions. In contrast, our framework maintains high stability when using either null text (“null text”) or Robot VAE latents (“Ours”). Ultimately, we opt for robot VAE latents as the primary conditioning signal, as they provide a dense, kinematically-informed representation that significantly enhances the fine-grained control over robot pointmap generation, surpassing the spatial ambiguity inherent in natural language. 


\paragraph{Depth map as the spatial representation.}
While our framework utilizes 4D pointmaps as the spatial representation, we evaluate the versatility of our model by training and testing on depth-map sequences. Leveraging our ST-v2 \cite{xiao2025spatialtracker} annotation pipeline, high-fidelity depth information is derived directly from our existing data without requiring additional effort. However, empirical results reported in \cref{tab:more_ablation} (“depth”) indicate that relying solely on depth maps leads to a performance degradation. We hypothesize that while depth captures distances from the camera, it lacks the explicit tri-coordinate geometric constraints ($X, Y, Z$) inherent in pointmaps. This weakens the model's ability to maintain precise metric-space control, particularly for off-axis manipulation tasks. 

\paragraph{Discussion on efficiency and memory usage.}
We provide a comparison of computational efficiency and peak memory usage against the leading 2D-based baseline, Ctrl-World \cite{guo2026ctrl}, and the 4D-based framework, TesserAct \cite{zhen2025tesseract}. 
All evaluations were conducted on a single NVIDIA A100 GPU to ensure parity. We also report the results of cross-domain evaluation, which is trained on DROID \cite{khazatsky2024droid} but tested on Bridge \cite{walke2023bridgedata}. All the baseline implementations use their official codes.
As summarized in \cref{tab:effi_cross}, although both TesserAct and Ctrl-World show better efficiency, they are limited in cross-domain ability.

Especially, Ctrl-World uses an auto-regressive architecture that offers superior throughput and a lower memory footprint. However, by relying on token-based action embeddings, it exhibits severe overfitting to the training dataset and lacks cross-embodiment capabilities due to the large gap between different embodiments' raw action spaces. Consequently, when deployed in unseen domains, it produces extremely inconsistent temporal frames (denoted by “(temp)” in \cref{tab:effi_cross}) and inferior visual quality due to the accumulated errors inherent in auto-regressive generation. As for TesserAct, it uses text instructions to represent action information, which lacks precise control and makes it fundamentally ineffective for the task-agnostic simulation.

In contrast, our framework prioritizes generative fidelity and structural consistency as a generalizable and robust foundation for 4D world simulation. We argue that for a foundation model, the priority must be the \textit{\textbf{reliability} and \textbf{generalizability} of the physical transitions}. We leave efficiency optimizations, such as knowledge distillation and model quantization, for future work.

\begin{table}[t]
\centering
\setlength{\tabcolsep}{6pt}
\caption{Comparison of the efficiency and cross-domain transfer. While the baseline methods offer better efficiency, their generalization abilities are extremely worse than our method, considering both the actual quality and the self-temporal consistency.}
\label{tab:effi_cross}
\resizebox{1.0\linewidth}{!}{
\begin{tabular}{l|cc||cc|cc|cc}
\toprule
Method & 
Time (min)  &  Memory (GB)  & 
\textbf{PSNR}$\uparrow$ & \textbf{PSNR (temp)$\uparrow$} & 
\textbf{FID$\downarrow$} & \textbf{FID (temp)$\downarrow$} & 
\textbf{CD-$L_1$$\downarrow$} & \textbf{CD-$L_1$ (temp)$\downarrow$}\\ 
\midrule
\textbf{Ours} & 
15  &  56  &
\textbf{19.23} & \textbf{36.95} & 
\textbf{29.3} & \textbf{15.4} & 
\textbf{0.0621} & \textbf{0.0068} \\ 
Ctrl-World \cite{guo2026ctrl}  & 
\textbf{2.5}  &  \textbf{15}  &
12.14 & 19.32 & 
38.5 & 49.5 & 
- & -\\
TesserAct \cite{zhen2025tesseract}  & 
 10 & 31 &
10.80 &  25.87 & 
41.8 & 28.7 & 
0.0928 & 0.0125\\
\bottomrule
\end{tabular}
}
\end{table}

\end{appendices}

%% file: main.bbl
\begin{thebibliography}{10}
\providecommand{\url}[1]{\texttt{#1}}
\providecommand{\urlprefix}{URL }
\providecommand{\doi}[1]{https://doi.org/#1}

\bibitem{abouchakra-embodiedgaussians}
Abou-Chakra, J., Rana, K., Dayoub, F., Suenderhauf, N.: Physically embodied gaussian splatting: A realtime correctable world model for robotics. In: CoRL (2024)

\bibitem{agarwal2025cosmos}
Agarwal, N., Ali, A., Bala, M., Balaji, Y., Barker, E., Cai, T., Chattopadhyay, P., Chen, Y., Cui, Y., Ding, Y., et~al.: Cosmos world foundation model platform for physical ai. arXiv preprint arXiv:2501.03575  (2025)

\bibitem{allen1983planning}
Allen, J.F., Koomen, J.A.: Planning using a temporal world model. In: IJCAI (1983)

\bibitem{assran2025v}
Assran, M., Bardes, A., Fan, D., Garrido, Q., Howes, R., Muckley, M., Rizvi, A., Roberts, C., Sinha, K., Zholus, A., et~al.: V-jepa 2: Self-supervised video models enable understanding, prediction and planning. arXiv preprint arXiv:2506.09985  (2025)

\bibitem{Qwen3-VL}
Bai, S., Cai, Y., Chen, R., Chen, K., Chen, X., Cheng, Z., Deng, L., Ding, W., Gao, C., Ge, C., Ge, W., Guo, Z., Huang, Q., Huang, J., Huang, F., Hui, B., Jiang, S., Li, Z., Li, M., Li, M., Li, K., Lin, Z., Lin, J., Liu, X., Liu, J., Liu, C., Liu, Y., Liu, D., Liu, S., Lu, D., Luo, R., Lv, C., Men, R., Meng, L., Ren, X., Ren, X., Song, S., Sun, Y., Tang, J., Tu, J., Wan, J., Wang, P., Wang, P., Wang, Q., Wang, Y., Xie, T., Xu, Y., Xu, H., Xu, J., Yang, Z., Yang, M., Yang, J., Yang, A., Yu, B., Zhang, F., Zhang, H., Zhang, X., Zheng, B., Zhong, H., Zhou, J., Zhou, F., Zhou, J., Zhu, Y., Zhu, K.: Qwen3-vl technical report. arXiv preprint arXiv:2511.21631  (2025)

\bibitem{rt12022arxiv}
Brohan, A., Brown, N., Carbajal, J., Chebotar, Y., Dabis, J., Finn, C., Gopalakrishnan, K., Hausman, K., Herzog, A., Hsu, J., Ibarz, J., Ichter, B., Irpan, A., Jackson, T., Jesmonth, S., Joshi, N., Julian, R., Kalashnikov, D., Kuang, Y., Leal, I., Lee, K.H., Levine, S., Lu, Y., Malla, U., Manjunath, D., Mordatch, I., Nachum, O., Parada, C., Peralta, J., Perez, E., Pertsch, K., Quiambao, J., Rao, K., Ryoo, M., Salazar, G., Sanketi, P., Sayed, K., Singh, J., Sontakke, S., Stone, A., Tan, C., Tran, H., Vanhoucke, V., Vega, S., Vuong, Q., Xia, F., Xiao, T., Xu, P., Xu, S., Yu, T., Zitkovich, B.: Rt-1: Robotics transformer for real-world control at scale. arXiv preprint arXiv:2212.06817  (2022)

\bibitem{brohan2022rt}
Brohan, A., Brown, N., Carbajal, J., Chebotar, Y., Dabis, J., Finn, C., Gopalakrishnan, K., Hausman, K., Herzog, A., Hsu, J., et~al.: Rt-1: Robotics transformer for real-world control at scale. arXiv preprint arXiv:2212.06817  (2022)

\bibitem{bruce2024genie}
Bruce, J., Dennis, M.D., Edwards, A., Parker-Holder, J., Shi, Y., Hughes, E., Lai, M., Mavalankar, A., Steigerwald, R., Apps, C., et~al.: Genie: Generative interactive environments. In: ICML (2024)

\bibitem{cen2025rynnvla}
Cen, J., Huang, S., Yuan, Y., Li, K., Yuan, H., Yu, C., Jiang, Y., Guo, J., Li, X., Luo, H., et~al.: Rynnvla-002: A unified vision-language-action and world model. arXiv preprint arXiv:2511.17502  (2025)

\bibitem{cen2025worldvla}
Cen, J., Yu, C., Yuan, H., Jiang, Y., Huang, S., Guo, J., Li, X., Song, Y., Luo, H., Wang, F., et~al.: Worldvla: Towards autoregressive action world model. arXiv preprint arXiv:2506.21539  (2025)

\bibitem{chang2025reconviagen}
Chang, J., Ye, C., Wu, Y., Chen, Y., Zhang, Y., Luo, Z., Li, C., Zhi, Y., Han, X.: Reconviagen: Towards accurate multi-view 3d object reconstruction via generation. arXiv preprint arXiv:2510.23306  (2025)

\bibitem{chefer2025videojam}
Chefer, H., Singer, U., Zohar, A., Kirstain, Y., Polyak, A., Taigman, Y., Wolf, L., Sheynin, S.: Videojam: Joint appearance-motion representations for enhanced motion generation in video models. In: ICML (2025)

\bibitem{chen2025world4omni}
Chen, H., Wang, B., Guo, J., Zhang, T., Hou, Y., Huang, X., Tie, C., Shao, L.: World4omni: a zero-shot framework from image generation world model to robotic manipulation. arXiv preprint arXiv:2506.23919  (2025)

\bibitem{chen2026bridgev2wbridgingvideogeneration}
Chen, Y., Li, P., Yang, J., He, K., Wu, X., Xu, Y., Wang, K., Liu, J., Liu, N., Huang, Y., Wang, L.: Bridgev2w: Bridging video generation models to embodied world models via embodiment masks. arXiv preprint arXiv:2602.03793  (2026)

\bibitem{chen20254dnex}
Chen, Z., Liu, T., Zhuo, L., Ren, J., Tao, Z., Zhu, H., Hong, F., Pan, L., Liu, Z.: 4dnex: Feed-forward 4d generative modeling made easy. arXiv preprint arXiv:2508.13154  (2025)

\bibitem{chen2024urdformer}
Chen, Z., Walsman, A., Memmel, M., Mo, K., Fang, A., Vemuri, K., Wu, A., Fox, D., Gupta, A.: Urdformer: A pipeline for constructing articulated simulation environments from real-world images. arXiv preprint arXiv:2405.11656  (2024)

\bibitem{chi2025diffusion}
Chi, C., Xu, Z., Feng, S., Cousineau, E., Du, Y., Burchfiel, B., Tedrake, R., Song, S.: Diffusion policy: Visuomotor policy learning via action diffusion. IJRR  (2025)

\bibitem{ebert2018visual}
Ebert, F., Finn, C., Dasari, S., Xie, A., Lee, A., Levine, S.: Visual foresight: Model-based deep reinforcement learning for vision-based robotic control. arXiv preprint arXiv:1812.00568  (2018)

\bibitem{erez2015simulation}
Erez, T., Tassa, Y., Todorov, E.: Simulation tools for model-based robotics: Comparison of bullet, havok, mujoco, ode and physx. In: ICRA (2015)

\bibitem{robomaster}
Fu, X., Wang, X., Liu, X., Bai, J., Xu, R., Wan, P., Zhang, D., Lin, D.: Learning video generation for robotic manipulation with collaborative trajectory control. In: ICLR (2026)

\bibitem{gao2026dreamdojo}
Gao, S., Liang, W., Zheng, K., Malik, A., Ye, S., Yu, S., Tseng, W.C., Dong, Y., Mo, K., Lin, C.H., et~al.: Dreamdojo: A generalist robot world model from large-scale human videos. arXiv preprint arXiv:2602.06949  (2026)

\bibitem{gu2025diffusion}
Gu, Z., Yan, R., Lu, J., Li, P., Dou, Z., Si, C., Dong, Z., Liu, Q., Lin, C., Liu, Z., et~al.: Diffusion as shader: 3d-aware video diffusion for versatile video generation control. In: SIGGRAPH (2025)

\bibitem{guo2026ctrl}
Guo, Y., Shi, L.X., Chen, J., Finn, C.: Ctrl-world: A controllable generative world model for robot manipulation. In: ICLR (2026)

\bibitem{ha2018world}
Ha, D., Schmidhuber, J.: World models. arXiv preprint arXiv:1803.10122  (2018)

\bibitem{han2025re}
Han, X., Liu, M., Chen, Y., Yu, J., Lyu, X., Tian, Y., Wang, B., Zhang, W., Pang, J.: Generating high-fidelity simulation data via 3d-photorealistic real-to-sim for robotic manipulation. arXiv preprint arXiv:2502.08645  (2025)

\bibitem{heusel2017gans}
Heusel, M., Ramsauer, H., Unterthiner, T., Nessler, B., Hochreiter, S.: Gans trained by a two time-scale update rule converge to a local nash equilibrium. In: NeurIPS (2017)

\bibitem{ho2020denoising}
Ho, J., Jain, A., Abbeel, P.: Denoising diffusion probabilistic models. In: NeurIPS (2020)

\bibitem{hong2022cogvideo}
Hong, W., Ding, M., Zheng, W., Liu, X., Tang, J.: Cogvideo: Large-scale pretraining for text-to-video generation via transformers. arXiv preprint arXiv:2205.15868  (2022)

\bibitem{hore2010image}
Hore, A., Ziou, D.: Image quality metrics: Psnr vs. ssim. In: ICPR (2010)

\bibitem{hu2022lora}
Hu, E.J., Shen, Y., Wallis, P., Allen-Zhu, Z., Li, Y., Wang, S., Wang, L., Chen, W., et~al.: Lora: Low-rank adaptation of large language models. In: ICLR (2022)

\bibitem{hu2025vggt4d}
Hu, Y., Cheng, C., Yu, S., Guo, X., Wang, H.: Vggt4d: Mining motion cues in visual geometry transformers for 4d scene reconstruction. arXiv preprint arxiv:2511.19971  (2025)

\bibitem{huang2025vipe}
Huang, J., Zhou, Q., Rabeti, H., Korovko, A., Ling, H., Ren, X., Shen, T., Gao, J., Slepichev, D., Lin, C.H., Ren, J., Xie, K., Biswas, J., Leal-Taixe, L., Fidler, S.: Vipe: Video pose engine for 3d geometric perception. In: NVIDIA Research Whitepapers arXiv:2508.10934 (2025)

\bibitem{huang2025particleformer}
Huang, S., Chen, Q., Zhang, X., Sun, J., Schwager, M.: Particleformer: A 3d point cloud world model for multi-object, multi-material robotic manipulation. In: CoRL (2025)

\bibitem{huang2026pointworld}
Huang, W., Chao, Y.W., Mousavian, A., Liu, M.Y., Fox, D., Mo, K., Fei-Fei, L.: Pointworld: Scaling 3d world models for in-the-wild robotic manipulation. arXiv preprint arXiv:2601.03782  (2026)

\bibitem{huang2025ladi}
Huang, Y., Zhang, J., Zou, S., Liu, X., Hu, R., Xu, K.: Ladi-wm: A latent diffusion-based world model for predictive manipulation. arXiv preprint arXiv:2505.11528  (2025)

\bibitem{jiang2025gsworld}
Jiang, G., Chang, H., Qiu, R.Z., Liang, Y., Ji, M., Zhu, J., Dong, Z., Zou, X., Wang, X.: Gsworld: Closed-loop photo-realistic simulation suite for robotic manipulation. arXiv preprint arXiv:2510.20813  (2025)

\bibitem{jiang2025enerverse}
Jiang, Y., Chen, S., Huang, S., Chen, L., Zhou, P., Liao, Y., He, X., Liu, C., Li, H., Yao, M., et~al.: Enerverse-ac: Envisioning embodied environments with action condition. arXiv preprint arXiv:2505.09723  (2025)

\bibitem{khazatsky2024droid}
Khazatsky, A., Pertsch, K., Nair, S., Balakrishna, A., Dasari, S., et~al.: {DROID}: A large-scale in-the-wild robot manipulation dataset. In: RSS 2024 Workshop: Data Generation for Robotics (2024)

\bibitem{kim24openvla}
Kim, M., Pertsch, K., Karamcheti, S., Xiao, T., Balakrishna, A., Nair, S., Rafailov, R., Foster, E., Lam, G., Sanketi, P., Vuong, Q., Kollar, T., Burchfiel, B., Tedrake, R., Sadigh, D., Levine, S., Liang, P., Finn, C.: Openvla: An open-source vision-language-action model. arXiv preprint arXiv:2406.09246  (2024)

\bibitem{li2024robogsim}
Li, X., Li, J., Zhang, Z., Zhang, R., Jia, F., Wang, T., Fan, H., Tseng, K.K., Wang, R.: Robogsim: A real2sim2real robotic gaussian splatting simulator. arXiv preprint arXiv:2411.11839  (2024)

\bibitem{li2024evaluating}
Li, X., Hsu, K., Gu, J., Mees, O., Pertsch, K., Walke, H.R., Fu, C., Lunawat, I., Sieh, I., Kirmani, S., Levine, S., Wu, J., Finn, C., Su, H., Vuong, Q., Xiao, T.: Evaluating real-world robot manipulation policies in simulation. In: CORL (2024)

\bibitem{li2025megasam}
Li, Z., Tucker, R., Cole, F., Wang, Q., Jin, L., Ye, V., Kanazawa, A., Holynski, A., Snavely, N.: {MegaSaM}: Accurate, fast and robust structure and motion from casual dynamic videos. In: CVPR (2025)

\bibitem{liang2025wonderland}
Liang, H., Cao, J., Goel, V., Qian, G., Korolev, S., Terzopoulos, D., Plataniotis, K.N., Tulyakov, S., Ren, J.: Wonderland: Navigating 3d scenes from a single image. In: CVPR (2025)

\bibitem{liao2026genie}
Liao, Y., Zhou, P., Huang, S., Yang, D., Chen, S., Jiang, Y., Hu, Y., Liu, S., Luo, J., Chen, L., YAN, S., Yao, M., Ren, G.: Genie envisioner: A unified world foundation platform for robotic manipulation. In: ICLR (2026)

\bibitem{libero}
Liu, B., Zhu, Y., Gao, C., Feng, Y., Liu, Q., Zhu, Y., Stone, P.: Libero: Benchmarking knowledge transfer for lifelong robot learning. In: NeurIPS (2023)

\bibitem{liu2025manipulation}
Liu, M., Zhu, Z., Han, X., Hu, P., Lin, H., Li, X., Chen, J., Xu, J., Yang, Y., Lin, Y., et~al.: Manipulation as in simulation: Enabling accurate geometry perception in robots. arXiv preprint arXiv:2509.02530  (2025)

\bibitem{liu2023grounding}
Liu, S., Zeng, Z., Ren, T., Li, F., Zhang, H., Yang, J., Li, C., Yang, J., Su, H., Zhu, J., et~al.: Grounding dino: Marrying dino with grounded pre-training for open-set object detection. arXiv preprint arXiv:2303.05499  (2023)

\bibitem{robo4dgen}
Liu, Z., Li, S., Cousineau, E., Feng, S., Burchfiel, B., Song, S.: Geometry-aware 4d video generation for robot manipulation. In: ICLR (2026)

\bibitem{lu2025gwm}
Lu, G., Jia, B., Li, P., Chen, Y., Wang, Z., Tang, Y., Huang, S.: Gwm: Towards scalable gaussian world models for robotic manipulation. In: ICCV (2025)

\bibitem{luo2025learning}
Luo, H., Lu, Z.: Learning video-conditioned policy on unlabelled data with joint embedding predictive transformer. In: ICLR (2025)

\bibitem{makoviychuk2021isaac}
Makoviychuk, V., Wawrzyniak, L., Guo, Y., Lu, M., Storey, K., Macklin, M., Hoeller, D., Rudin, N., Allshire, A., Handa, A., et~al.: Isaac gym: High performance gpu-based physics simulation for robot learning. arXiv preprint arXiv:2108.10470  (2021)

\bibitem{mittal2023orbit}
Mittal, M., Yu, C., Yu, Q., Liu, J., Rudin, N., Hoeller, D., Yuan, J.L., Singh, R., Guo, Y., Mazhar, H., et~al.: Orbit: A unified simulation framework for interactive robot learning environments. IRAL  (2023)

\bibitem{robotwin}
Mu, Y., Chen, T., Chen, Z., Peng, S., Lan, Z., Gao, Z., Liang, Z., Yu, Q., Zou, Y., Xu, M., Lin, L., Xie, Z., Ding, M., Luo, P.: Robotwin: Dual-arm robot benchmark with generative digital twins. In: CVPR (2025)

\bibitem{o2024open}
O’Neill, A., Rehman, A., Maddukuri, A., Gupta, A., Padalkar, A., Lee, A., Pooley, A., Gupta, A., Mandlekar, A., Jain, A., et~al.: Open x-embodiment: Robotic learning datasets and rt-x models: Open x-embodiment collaboration 0. In: ICRA (2024)

\bibitem{pashevich2019learning}
Pashevich, A., Strudel, R., Kalevatykh, I., Laptev, I., Schmid, C.: Learning to augment synthetic images for sim2real policy transfer. In: IROS (2019)

\bibitem{dit}
Peebles, W., Xie, S.: Scalable diffusion models with transformers. In: ICCV (2023)

\bibitem{ranftl2020towards}
Ranftl, R., Lasinger, K., Hafner, D., Schindler, K., Koltun, V.: Towards robust monocular depth estimation: Mixing datasets for zero-shot cross-dataset transfer. TPAMI  (2020)

\bibitem{ravi2024sam2}
Ravi, N., Gabeur, V., Hu, Y.T., Hu, R., Ryali, C., Ma, T., Khedr, H., R{\"a}dle, R., Rolland, C., Gustafson, L., Mintun, E., Pan, J., Alwala, K.V., Carion, N., Wu, C.Y., Girshick, R., Doll{\'a}r, P., Feichtenhofer, C.: Sam 2: Segment anything in images and videos. arXiv preprint arXiv:2408.00714  (2024), \url{https://arxiv.org/abs/2408.00714}

\bibitem{ren2025gen3c}
Ren, X., Shen, T., Huang, J., Ling, H., Lu, Y., Nimier-David, M., M{\"u}ller, T., Keller, A., Fidler, S., Gao, J.: Gen3c: 3d-informed world-consistent video generation with precise camera control. In: CVPR (2025)

\bibitem{rombach2022high}
Rombach, R., Blattmann, A., Lorenz, D., Esser, P., Ommer, B.: { High-Resolution Image Synthesis with Latent Diffusion Models }. In: CVPR (2022)

\bibitem{schonberger2016structure}
Schonberger, J.L., Frahm, J.M.: Structure-from-motion revisited. In: CVPR (2016)

\bibitem{schonberger2016pixelwise}
Sch{\"o}nberger, J.L., Zheng, E., Frahm, J.M., Pollefeys, M.: Pixelwise view selection for unstructured multi-view stereo. In: ECCV (2016)

\bibitem{song2019generative}
Song, Y., Ermon, S.: Generative modeling by estimating gradients of the data distribution. In: NeurIPS (2019)

\bibitem{su2024roformer}
Su, J., Ahmed, M., Lu, Y., Pan, S., Bo, W., Liu, Y.: Roformer: Enhanced transformer with rotary position embedding. Neurocomput.  (2024)

\bibitem{Aether}
Team, A., Zhu, H., Wang, Y., Zhou, J., Chang, W., Zhou, Y., Li, Z., Chen, J., Shen, C., Pang, J., He, T.: Aether: Geometric-aware unified world modeling. In: ICCV (2025)

\bibitem{team2025evaluating}
Team, G.R., Choromanski, K., Devin, C., Du, Y., Dwibedi, D., Gao, R., Jindal, A., Kipf, T., Kirmani, S., Leal, I., et~al.: Evaluating gemini robotics policies in a veo world simulator. arXiv preprint arXiv:2512.10675  (2025)

\bibitem{todorov2012mujoco}
Todorov, E., Erez, T., Tassa, Y.: Mujoco: A physics engine for model-based control. In: IROS (2012)

\bibitem{walke2023bridgedata}
Walke, H., Black, K., Lee, A., Kim, M.J., Du, M., Zheng, C., Zhao, T., Hansen-Estruch, P., Vuong, Q., He, A., Myers, V., Fang, K., Finn, C., Levine, S.: Bridgedata v2: A dataset for robot learning at scale. In: CoRL (2023)

\bibitem{wan2025wan}
Wan, T., Wang, A., Ai, B., Wen, B., Mao, C., Xie, C.W., Chen, D., Yu, F., Zhao, H., Yang, J., et~al.: Wan: Open and advanced large-scale video generative models. arXiv preprint arXiv:2503.20314  (2025)

\bibitem{wang2025vggt}
Wang, J., Chen, M., Karaev, N., Vedaldi, A., Rupprecht, C., Novotny, D.: Vggt: Visual geometry grounded transformer. In: CVPR (2025)

\bibitem{VAP_2025_ICCV}
Wang, Y., Wen, C., Guo, H., Peng, S., Qin, M., Bao, H., Zhou, X., Hu, R.: Precise action-to-video generation through visual action prompts. In: ICCV (2025)

\bibitem{wang2004image}
Wang, Z., Bovik, A.C., Sheikh, H.R., Simoncelli, E.P.: Image quality assessment: from error visibility to structural similarity. TIP  (2004)

\bibitem{wu2024ivideogpt}
Wu, J., Yin, S., Feng, N., He, X., Li, D., Hao, J., Long, M.: ivideogpt: Interactive videogpts are scalable world models. In: NeurIPS (2024)

\bibitem{xiang2020sapien}
Xiang, F., Qin, Y., Mo, K., Xia, Y., Zhu, H., Liu, F., Liu, M., Jiang, H., Yuan, Y., Wang, H., et~al.: Sapien: A simulated part-based interactive environment. In: CVPR (2020)

\bibitem{xiao2025spatialtracker}
Xiao, Y., Wang, J., Xue, N., Karaev, N., Makarov, I., Kang, B., Zhu, X., Bao, H., Shen, Y., Zhou, X.: Spatialtrackerv2: 3d point tracking made easy. In: ICCV (2025)

\bibitem{xu2025stable}
Xu, M., Ye, C., Liu, H., Wu, Y., Chang, J., Han, X.: Stable-sim2real: Exploring simulation of real-captured 3d data with two-stage depth diffusion. In: ICCV (2025)

\bibitem{yang2025roboenvision}
Yang, L., Bai, Y., Eskandar, G., Shen, F., Altillawi, M., Chen, D., Majumder, S., Liu, Z., Kutyniok, G., Valada, A.: Roboenvision: A long-horizon video generation model for multi-task robot manipulation. In: IROS (2025)

\bibitem{yang2024learning}
Yang, S., Du, Y., Ghasemipour, S.K.S., Tompson, J., Kaelbling, L.P., Schuurmans, D., Abbeel, P.: Learning interactive real-world simulators. In: ICLR (2024)

\bibitem{yang2024hunyuan3d}
Yang, X., Shi, H., Zhang, B., Yang, F., Wang, J., Zhao, H., Liu, X., Wang, X., Lin, Q., Yu, J., Wang, L., Chen, Z., Liu, S., Liu, Y., Yang, Y., Wang, D., Jiang, J., Guo, C.: Tencent hunyuan3d-1.0: A unified framework for text-to-3d and image-to-3d generation. arXiv preprint arXiv:2411.02293  (2024)

\bibitem{yang2026orv}
Yang, X., Li, B., Xu, S., Wang, N., Ye, C., Zhaoxi, C., Qin, M., Yikang, D., Jin, X., Zhao, H., Zhao, H.: Orv: 4d occupancy-centric robot video generation. In: CVPR (2026)

\bibitem{ye2025anchordream}
Ye, J., Xue, R., Van~Hoorick, B., Tokmakov, P., Irshad, M.Z., Wang, Y., Guizilini, V.: Anchordream: Repurposing video diffusion for embodiment-aware robot data synthesis. arXiv preprint arXiv:2512.11797  (2025)

\bibitem{yu2025wonderworld}
Yu, H.X., Duan, H., Herrmann, C., Freeman, W.T., Wu, J.: Wonderworld: Interactive 3d scene generation from a single image. In: CVPR (2025)

\bibitem{Yu_2025_ICCV}
Yu, M., Hu, W., Xing, J., Shan, Y.: Trajectorycrafter: Redirecting camera trajectory for monocular videos via diffusion models. In: ICCV (2025)

\bibitem{zhang2025imowm}
Zhang, C., Wu, Z., Lu, G., Tang, Y., Wang, Z.: imowm: Taming interactive multi-modal world model for robotic manipulation. arXiv preprint arXiv:2510.09036  (2025)

\bibitem{zhang2025show}
Zhang, D.J., Wu, J.Z., Liu, J.W., Zhao, R., Ran, L., Gu, Y., Gao, D., Shou, M.Z.: Show-1: Marrying pixel and latent diffusion models for text-to-video generation. IJCV  (2025)

\bibitem{zhang2024monst3r}
Zhang, J., Herrmann, C., Hur, J., Jampani, V., Darrell, T., Cole, F., Sun, D., Yang, M.H.: Monst3r: A simple approach for estimating geometry in the presence of motion. arXiv preprint arxiv:2410.03825  (2024)

\bibitem{zhang2023adding}
Zhang, L., Rao, A., Agrawala, M.: Adding conditional control to text-to-image diffusion models. In: ICCV (2023)

\bibitem{zhang2018unreasonable}
Zhang, R., Isola, P., Efros, A.A., Shechtman, E., Wang, O.: The unreasonable effectiveness of deep features as a perceptual metric. In: CVPR (2018)

\bibitem{zhen2025tesseract}
Zhen, H., Sun, Q., Zhang, H., Li, J., Zhou, S., Du, Y., Gan, C.: Tesseract: learning 4d embodied world models. In: ICCV (2025)

\bibitem{robodreamer}
Zhou, S., Du, Y., Chen, J., Li, Y., Yeung, D.Y., Gan, C.: Robodreamer: learning compositional world models for robot imagination. In: ICML (2024)

\bibitem{zhu2025irasim}
Zhu, F., Wu, H., Guo, S., Liu, Y., Cheang, C., Kong, T.: Irasim: Learning interactive real-robot action simulators. In: ICCV (2025)

\end{thebibliography}
